\documentclass[review]{elsarticle}

\usepackage{lineno,hyperref}
\modulolinenumbers[5]

\usepackage{algorithm}
\usepackage{algorithmic}

\newcommand{\tabincell}[2]{\begin{tabular}{@{}#1@{}}#2\end{tabular}}

\usepackage{amsmath}
\usepackage{array}
\usepackage{multirow}

\usepackage{url}

\journal{Pattern Recognition}

\begin{document}

\begin{frontmatter}

\title{Towards Interpretable and Robust Hand Detection via Pixel-wise Prediction}




\author[1,4]{Dan Liu}

\author[2,5]{Libo Zhang\corref{mycorrespondingauthor}}
\cortext[mycorrespondingauthor]{Corresponding author}
\ead{libo@iscas.ac.cn} 

\author[1,4]{Tiejian Luo}

\author[3]{Lili Tao}
\author[2]{Yanjun Wu}

\address[1]{University of Chinese Academy of Sciences, China 100049}
\address[2]{Institute of Software Chinese Academy of Sciences,  China 100190}
\address[3]{University of the West of England, Bristol BS16 1QY, U.K.}
\fntext[4]{Dan Liu and Tiejian Luo were contributed equally and should be considered as co-first authors. Models and code are available at https://isrc.iscas.ac.cn/gitlab/research/pr2020-phdn.}
\fntext[5]{This work was supported by the National Natural Science Foundation of China, Grant No. 61807033, the Key Research Program of Frontier Sciences, CAS, Grant No. ZDBS-LY-JSC038. Libo Zhang was supported by Youth Innovation Promotion Association, CAS, and Outstanding Youth Scientist Project of ISCAS.}

\begin{abstract}
The lack of interpretability of existing CNN-based hand detection methods makes it difficult to understand the rationale behind their predictions. In this paper, we propose a novel neural network model, which introduces interpretability into hand detection for the first time. The main improvements include: (1) Detect hands at pixel level to explain what pixels are the basis for its decision and improve transparency of the model. (2) The explainable Highlight Feature Fusion block highlights distinctive features among multiple layers and learns discriminative ones to gain robust performance. (3) We introduce a transparent representation, the rotation map, to learn rotation features instead of complex and non-transparent rotation and derotation layers. (4) Auxiliary supervision accelerates the training process, which saves more than 10 hours in our experiments. Experimental results on the VIVA and Oxford hand detection and tracking datasets show competitive accuracy of our method compared with state-of-the-art methods with higher speed.
\end{abstract}


\end{frontmatter}


\section{Introduction}
%
%
%
%
Deep neural networks are widely adopted in many fields of study, \textit{e.g.}, computer vision and natural language processing, and achieve state-of-the-art results. However, as their inner workings are not transparent, the correctness and objectivity of the predicting results cannot be guaranteed and thus limit their development in industry. In recent years, some researchers have begun to explore interpretable deep leaning methods. \cite{zhang2017mdnet} focuses on network interpretability in medical image diagnosis. \cite{montavon2017explaining} decomposes output into contributions of its input features to interpret the image classification network. There is also a clear need to develop an interpretable neural network in driving monitoring as the predicting results will directly affect the safety of drivers, passengers, and pedestrians. In this paper, we present a highly interpretable neural network to detect hands in images, which is a basic task in driving monitoring.

Hand detection in natural scenes plays an important role in virtual reality, human-computer interaction, driving monitoring~\cite{7, 36}. It is a critical and primary task for higher-level tasks such as hand tracking, gesture recognition, human activity understanding. Particularly, accurately detecting hand is a vital part in monitoring driving behavior~\cite{36, 37}. Detecting hands in images is a challenging task. The illumination conditions, occlusion, and color/shape similarity will bring great difficulties to hand detection. Moreover, hands are highly deformable objects, which hard to detect due to their variability and flexibility. Hands are not always shown in an upright position in images, so the rotation angle needs to be considered to locate the hand in images more accurately.

The problem of hand detection has been studied for years. Traditional methods extract features such as skin-related features~\cite{3}, hand shape and background, Histograms of Oriented Gradients (HOG)~\cite{2} to build feature vector for each sample. Then these vectors are used to train classifiers such as SVM~\cite{12}. Although the hand-crafted features have clear meanings and are easy to understand, they are too limited to meet the requirements for the accuracy of hand detection in the real world. With the increasing influence of Convolutional Neural Networks (CNNs) in the field of computer vision, many CNN-based object detection methods have emerged, Region-Based Convolutional Networks(R-CNNs)~\cite{4}, Single Shot MultiBox Detector (SSD)~\cite{6}, for example. Inspired by these advances, many CNN-based methods have been proposed to deal with hand detection. Features are extracted automatically by designed CNNs from the original images~\cite{8,23} or the region proposals~\cite{7} and then used to locate the hands in original images. In order to extract as many effective features as possible to detect hand more accurately, the network structure is always very complicated and therefore has a heavy computational burden. This limits its value in practical applications such as monitoring driving behavior and sign language recognition. The deep CNNs are used as black-boxes in the existing methods. Different from hand-crafted features, it is difficult to know the meaning of features extracted by CNNs. As a result, the stability and robustness of these methods cannot be guaranteed.

%
\begin{figure}[!t]
	\centering
	\includegraphics[width=3.2in]{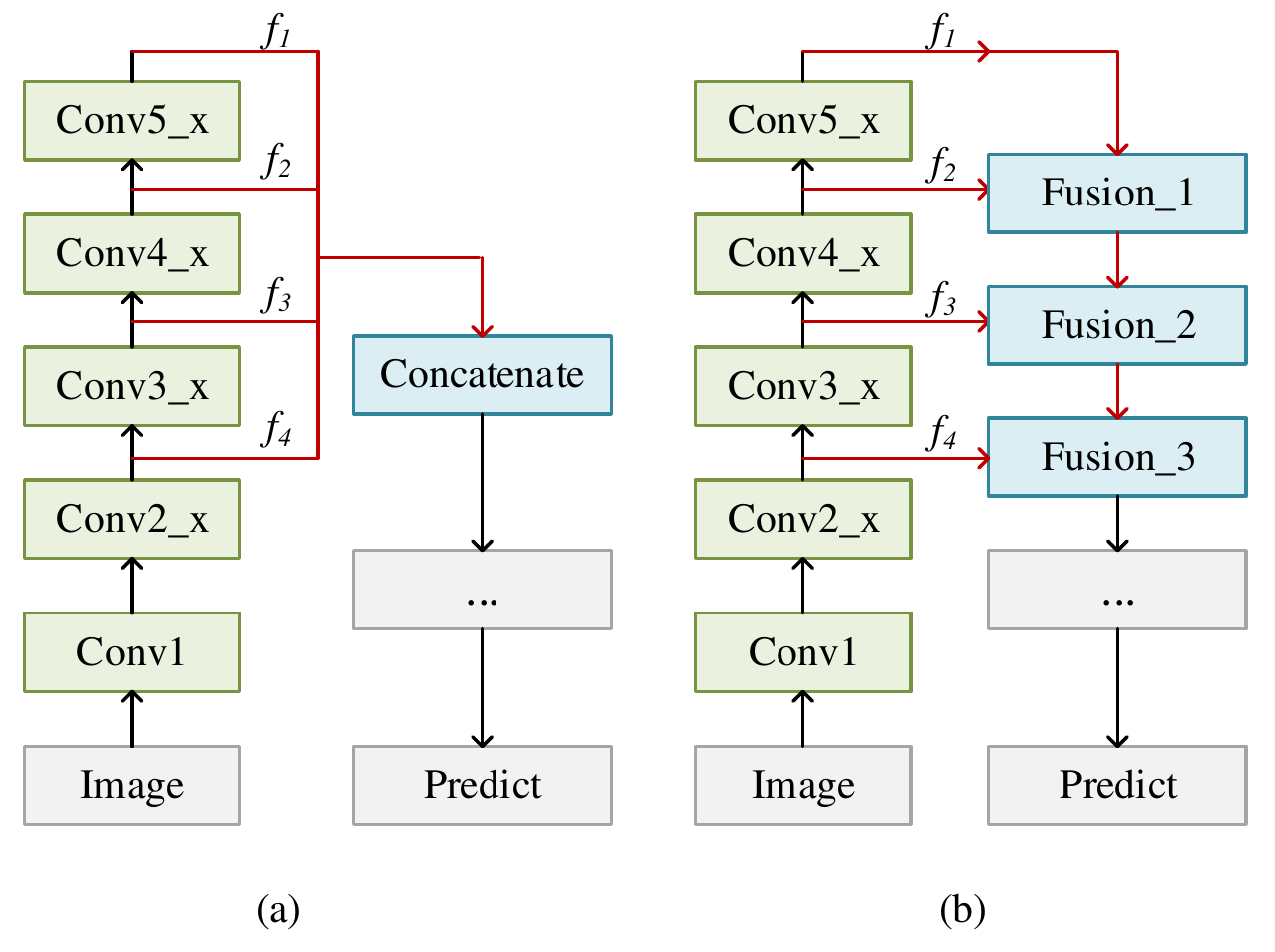}
	\caption{Different connection modes of multi-scale features. (a) Serial mode. (b) Cascade mode.}
	\label{connection_mode}
\end{figure}


In view of the issues mentioned above, we propose an interpretable framework, Pixel-wise Hand Detection Network (PHDN), to detect hands more efficiently. The proposed method achieves better performance with faster computational speed. An explainable module named Highlight Feature Fusion (HFF) block is developed to get more discriminative features. With HFF block, PHDN performs effectively and stably in different image contexts. To the best of our knowledge, this is the first time to give reasonable explanations of learned features in the hand detection procedure. Popular deep convolutional neural networks VGG16~\cite{13} or ResNet50~\cite{10} is adopted as a backbone network in PHDN. The HFF block makes full use of multi-scale features by weighting the lower-level features with the higher-level features. In this way, the discriminative features, namely the effective ones for locating the hand, are highlighted in the detection procedure. Each HFF block fuses features from two layers. It first weights the lower-level features by the last higher-level feature maps and then fuses the features by convolution operations. Several HFF blocks are connected in cascade mode (see Fig.~\ref{connection_mode}(b)) to iteratively fuse multi-scale features, which greatly reduces computational overhead and saves time compared to the serial connection (see Fig.~\ref{connection_mode}(a)). As PHDN makes hand region predictions with multi-scale features, it is more robust to hands of different sizes. In other words, our model is scale-invariance. 

As for the rotated hand detection, adding additional rotation and derotation layers \cite{9} makes the network more complicated and thus increases the computational burden and time overhead. We propose the rotation map and the distance map to store the rotation angle and the geometry information of the hand region respectively, which handles the rotation hands without increasing complexity of the network and learns more interpretable representations of angles by recording angles of pixels directly. 

In the training process, we add supervision to each HFF block. Deep supervision to the hidden layers makes the learned features more discriminative and robust, and thus the performance of the detector is better. The auxiliary losses accelerate the convergence of training in a simple and direct way compared with \cite{60}, which accelerates training by constraining the input weight of each neuron with zero mean and unit norm. 

Existing detection methods make predictions for grid cells~\cite{25} or default boxes~\cite{6}, which need to seek appropriate anchor scales. Alternatively, we predict hand regions at pixel resolution to avoid the adverse effects of improper anchor scales settings, for which we name our model as Pixel-wise Hand Detection Network. Detecting hands at pixel level also explains what pixels are the basis for its decision, which improves transparency of the model. The hand regions predicted by PHDN are filtered by the Non-Maximum Suppression (NMS) to yield the final detection results.


To evaluate our model, experiments are conducted on two authentic and publicly accessible hand detection datasets, the VIVA hand detection dataset~\cite{11} and the Oxford hand detection dataset~\cite{12}. Compared with the state-of-the-art methods, our model achieves competitive Average Precision (AP) and Average Recall (AR) on VIVA dataset with 4.23 times faster detecting speed, and obtains 5.5\% AP improvement on Oxford dataset. Furthermore, we test the PHDN with the hand tracking task on VIVA hand tracking dataset~\cite{42}, which is a higher application scenario of hand detection. We try three tracking-by-detection methods: SORT tracker~\cite{39}, deep SORT tracker~\cite{40} and IOU tracker~\cite{41}, where the PHDN acts as a detector. Experimental results show that using any of the aforementioned tracking algorithms based on our detector can achieve better results than existing methods. It indicates that PHDN is robust and practicable as the detector performance plays a crucial role in tracking-by-detection multiple object tracking methods.

Part of the work has been introduced in \cite{57}. The extensions made in this article compared to \cite{57} are as follows: (1) We analyze the interpretability of our model by visualizing the features extracted by HFF block to interpret our model. It shows the mechanism of internal layers and demonstrates how our method outperforms the others. (2) We integrate our detector with the popular trackers to track hands in videos and achieve state-of-the-art results on the authoritative VIVA hand tracking challenge dataset \cite{42}. (3) We give a more detailed description of our model including related work in hand detection and multiple hand tracking in vehicles, network architecture, feature fusion processing, loss functions and the settings and results of conducted experiments. 
 
The main contributions of this paper are in four folds:
\begin{itemize}
	\item We give insight to the interpretability of the hand detection network for the first time. Reasonable explanations for the feature activated in hand detection procedure and the discriminative features learned by HFF block are first given. The proposed Pixel-wise Hand Detection Network predicts hand regions at pixel resolution rather than grid cells or default boxes. It gets rid of the adverse effects of inappropriate anchor scales and can detect different sizes of hands by fusing multi-scale features with the cascaded HFF blocks.  
	\item The rotation map is designed to predict hand rotation angles precisely. It learns and represents the angles in an interpretable way with less computational cost.
	\item Auxiliary losses are added to provide supervision to hidden layers of the network, leading to faster convergence of the training and higher precision.
	\item Experiments on VIVA and Oxford hand detection datasets show that PHDN achieves competitive performance compared with the state-of-the-art methods. Evaluated on the VIVA hand tracking dataset, tracking-by-detection trackers such as SORT tracker, deep SORT tracker and IOU tracker with the PHDN detector outperform the existing hand tracking methods.
\end{itemize}

The remainder of this paper is organized as follows. In Section~\ref{related_work}, we review the related work in the field. Section~\ref{proposed_framework} gives a detailed description of the proposed method. Section~\ref{experiment} introduces the datasets and experimental setup, reports and analyzes the results. Finally, concluding remarks are presented in Section~\ref{conclusion}.

\begin{figure}[!t]
	\centering
	\includegraphics[width=3.8in]{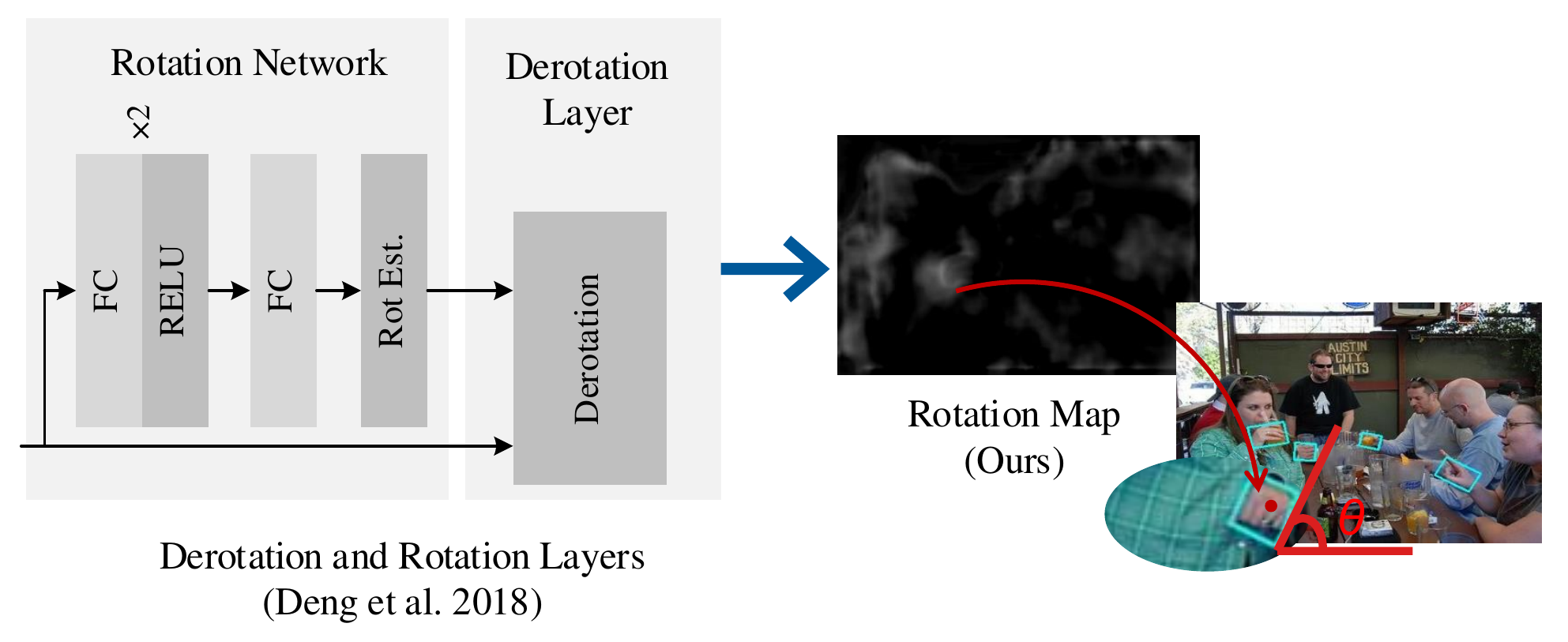}
	\caption{Novel and transparent representation of the rotation angle. We use the rotation map to store the rotation angle instead of adding rotation and derotation layers \cite{9} to networks.}
	\label{rotation_map}
\end{figure}

\section{Related Work} \label{related_work}
\subsection{Hand Detection}
Current hand detection methods can be divided into two categories. One is based on the hand-crafted structured features, such as color, shape and so on. The other is based on features extracted by CNNs. The methods based on hand-crafted features have strong interpretability, but the detection performance is poor due to the limitations of features. On the contrary, CNNs-based methods tend to have good performance but poor interpretability.
\subsubsection{Human-interpretable Features Based Methods}
Hand detection methods that use human crafted features usually propose hand regions using features like skin color, hand shape, Histograms of Oriented Gradients (HOG) \cite{44}. These features have specific meanings and are easy to understand. Then the features are used to train a classifier, such as Support Vector Machine (SVM)~\cite{12}, to generate the final detection results. \cite{32} uses the skin and hand shape features to detect hands from images. Skin areas are extracted first using a skin detector and the hands are separated out using hand contour comparison. However, it may be confusing when distinguishing between face and fist since their contours are similar. \cite{12} generates hand region proposals using a hand shape detector, a context-based detector and a skin-based detector. Then a SVM classifier, with the score vectors built by the three detectors as input, is trained to classify the hand and non-hand regions. To enhance the robustness of hand detection in cluttered background, \cite{33} proposes three new features based on HOG, Local Binary Patterns (LBP) and Local Trinary Patterns (LTP) descriptors to train classifiers, but it does not perform well if the image is low resolution and it cannot handle well with occlusion. \cite{2} trains a SVM classifier with the HOG features, and extends it with a Dynamic Bayesian Network for better performance. Due to the limitation of hand-crafted features, these methods are not robust to the change of illumination, background and hand shape. Moreover, the non-end-to-end optimization process is time-consuming and the performance is often suboptimal.

\subsubsection{Non-transparent CNNs Based Methods}
Inspired by the progress of Convolutional Neural Networks (CNNs), many hand detection methods proposed recently are based on CNNs.
\cite{7} presents a lightweight hand proposal generation approach, of which a CNN-based method is used to disambiguate hands in complex egocentric interactions. Context information, such as hand shapes and locations, can be seen as prior knowledge, and they can be used to train a hand detector~\cite{24}. However, it is no doubt that additional context cues over-complicates the image preprocessing step. Inspired by these, \cite{8} first generates hand region proposals with the Fully Convolutional Network (FCN)~\cite{15} and then fuses multi-scale features extracted from FCN into a large feature map to make final predictions, as a result of which the convolution operations are time-consuming in the later steps. Similarly, \cite{23} concatenates the multi-scale feature maps from the last three pooling layers into a large feature map. Although different receptive fields are taken into consideration, simple concatenation of feature maps results in high computational cost.

In contrast to human-crafted features, the features extracted by CNNs are not interpretable and thus the rationality and validity of the model are difficult to verify. In order to provide interpretability to CNN-based hand detection models, we detect hands at pixel level. For any pixel in the image, we predict whether it belongs to a hand and the bounding box of the hand. In this way, we can know the basis for the model to make predictions. Under the fact that the high-level feature maps reflect the global features while the low-level feature maps contain more local information, the feature maps from different scales are weighted before merged so that the features from multiple scales can complement each other in the subsequent process. In view of the heavy computational burden caused by the fusion of multi-scale information, our model fuses multi-scale features iteratively rather than simultaneously.

Another issue of hand detection is to handle the rotation. Hands are rarely shown in upright positions in images. To accurately detect hands and estimate their poses, \cite{9} designs a rotation network to predict the rotation angle of region proposals and a derotation layer to obtain axis-aligned rotating feature maps (see Fig.~\ref{rotation_map}). However, the method is of great complexity as it includes two components for rotation, a shared network for learning features and a detection network for the classification task. It is also hard to find out what the rotation and derotation layers really learn. To handle rotated hand samples more effectively, we develop the rotation map to replace the complex rotation and derotation layers, as shown in Fig.~\ref{rotation_map}. It is also more interpretable as each pixel value represents the rotation angle directly. The results on the Oxford hand detection dataset show that the rotation map brings a significant increase (about 0.30) in AP compared to using only the distance maps.

\subsection{Multiple Hand Tracking in Vehicles}
Tracking hands in the vehicle cabin is important for monitoring driving behavior and research in intelligent vehicles. Although hand tracking has been studied since the last century, there are few studies on tracking multiple hands simultaneously in naturalistic driving conditions. To the best of our knowledge, only \cite{37} has given the research results on multiple hand tracking so far. \cite{37} proposes a tracking-by-detection method, where each video frame is processed by the detector first and then integrates with a tracker to provide individual tracks online. The ACF detector~\cite{52} is used to generate hand detection results and the data association is performed using a bipartite matching algorithm. It reports the tracking results on the VIVA hand tracking dataset. 
To investigate the performance of our model in hand tracking, we apply PHDN to SORT tracker~\cite{39}, deep SORT tracker~\cite{40}, IOU tracker~\cite{41}. SORT tracker and deep SORT tracker are online tracking methods, where only the current and previous frames are visible to the tracker. SORT tracker performs Kalman filtering in image space and uses the Hungarian method to associate detections across frames in a video sequence. Deep SORT tracker is developed for the many identity switches in SORT tracker. It adopts a novel association metric with more motion and appearance information compared to the IOU distance used in SORT tracker. The reported results show the deep SORT tracker has fewer identity switches than the SORT tracker. IOU tracker is an offline tracking method that can generate trajectories with all observations in the video. It associates the detection with the highest IOU to the last detection in previous frames to extend a trajectory. It can run at 100K fps as its complexity is very low. The tracking performance depends largely on the detector. Therefore, we conduct experiments on the VIVA hand tracking dataset with our detector and we use three trackers to evaluate our model in the practical tracking task.

\begin{figure}[!t]
	\centering
	\includegraphics[width=0.87\textwidth]{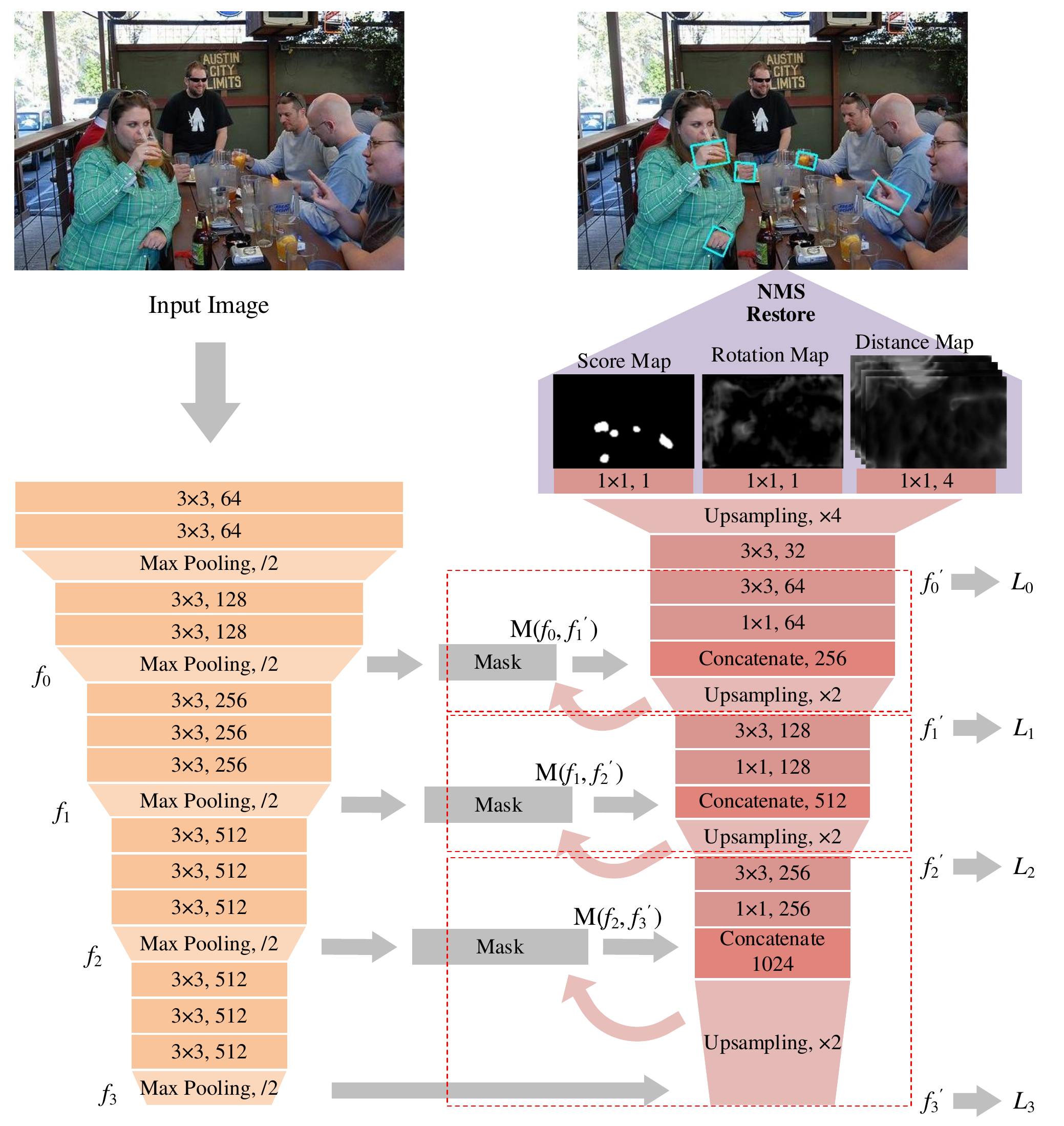}%
	\caption{PHDN architecture with VGG16 as the backbone. The left is feature extracting stem, and the right is feature fusion branch and the output layers. Highlight Feature Fusion (HFF) block is marked with red dotted rectangle.}
	\label{net_arch}
\end{figure}

\section{Interpretable Pixel-wise Hand Detection Network} \label{proposed_framework}
The PHDN architecture is illustrated in Fig.~\ref{net_arch}. To show our model more clearly, only the VGG16 backbone is presented in the figure for its simpler structure compared with ResNet50. The feature maps from four different scales extracted by the VGG16 extractor or ResNet extractor are fused iteratively in the cascaded HFF blocks. The final feature maps, containing multi-scale information, are upsampled and convoluted to get the score map, the rotation map and the distance map. With the three kinds of maps, we can restore the hand bounding boxes and filter them by the NMS to generate the final hand regions. In the following, we describe the pipeline in detail and construct the loss function for the training.
\subsection{Feature Extraction}
We try two popular deep convolutional networks, \textit{i.e.}, VGG16 and ResNet50, to extract features from the images. The pre-trained model on the ImageNet dataset~\cite{20} is used in our study. Feature maps from four layers are selected for the feature fusion module. For VGG16, we adopt the feature maps from \textit{pooling-2} to \textit{pooling-5}. Similarly, the outputs of \textit{conv2\_1}, \textit{conv3\_1}, \textit{conv4\_1} and \textit{conv5\_1} are extracted in ResNet50. The feature maps extracted from VGG16 or ResNet50 are $(\frac{1}{4})^2, (\frac{1}{8})^2, (\frac{1}{16})^2, (\frac{1}{32})^2$ the size of input images, and represent information of different sizes of receptive fields. 
\begin{algorithm}[H]
	\caption{Feature Fusion Procedure}
	\begin{algorithmic}[1]
		\label{alg:HFF}
		\REQUIRE~~\\
		Feature maps extracted by VGG16 or Resnet50, $f_s, s \in \{0,1,2,3\}$;\\
		Channels of fused feature maps, $c_s, s \in \{0,1,2,3\}$;
		\ENSURE~~\\
		Fused feature maps, $f'_s, s \in \{0,1,2,3\}$; 
		\STATE $f'_3 = f_3$;
		\FOR{$s$ from $2$ to $0$}
		\STATE $u_{s+1}=Upsampling(f'_{s+1})$;
		\STATE $masked=f_s * (1-Convolution(u_{s+1}, 1\times1))$;
		\label{code:mask}
		\STATE $Concate = Concatenate(masked, u_{s+1})$; 
		\STATE $Conv1 = Convolution(Concate, 1\times1, c_s)$;
		\STATE $Conv2 = Convolution(Concate, 3\times3, c_s)$;
		\STATE $f'_s=Conv2$
		\ENDFOR
		
		\RETURN $f'_s, s \in \{0,1,2,3\}$;
	\end{algorithmic}
\end{algorithm}

\subsection{Visually Interpretable and Robust Feature Fusion} \label{s_fusion}
The size of hands varies greatly in different images or even the same image. The larger hand detection needs more global information. It is known that the higher the level of feature maps, the more global the information is presented. Hence multi-scale feature maps should be merged to detect different sizes of hands. We propose to fuse the feature maps from multiple layers in an iterative way to reduce the computational cost, which can be achieved by cascaded feature fusion blocks as shown in Fig.~\ref{connection_mode}(b) To reduce the interference of useless features and learn more discriminative features, we develop the Highlight Feature Fusion (HFF) block to fuse the features from different scales. Fig.~\ref{net_arch} displays three cascaded HFF blocks, which are marked with red dotted rectangles. The cascaded HFF blocks operate the fusion as Algorithm \ref{alg:HFF}.

We generate a mask with the higher-level feature maps to filter the common features in the current level feature maps, which formulated as Line~\ref{code:mask} above and $*$ denotes element-wise multiplication. Masking $f_{s}$ with the complementary feature maps of $u_{s+1}$ can highlight the fine-grained distinctive information contained in $f_{s}$ that $u_{s+1}$ may not have. $Conv1$ is the result of conducting a $1\times1$ convolution on the concatenated feature maps. It is designed to reduce the output channels and thus lessen the computational burden. Then a $3\times3$ convolution is operated to further fuse the features of multiple scales. To investigate the effect of the mask, we remove the mask operation and concatenate $f_{s}$ and $u_{s+1}$ directly as a Base Feature Fusion (BFF) block in our experiments.

We visualize features extracted by HFF block and BFF block to interpret the robustness and effectiveness of HFF block in Section \ref{interpret}. 

\begin{figure}[!t]
	\centering
	\includegraphics[width=0.96\textwidth]{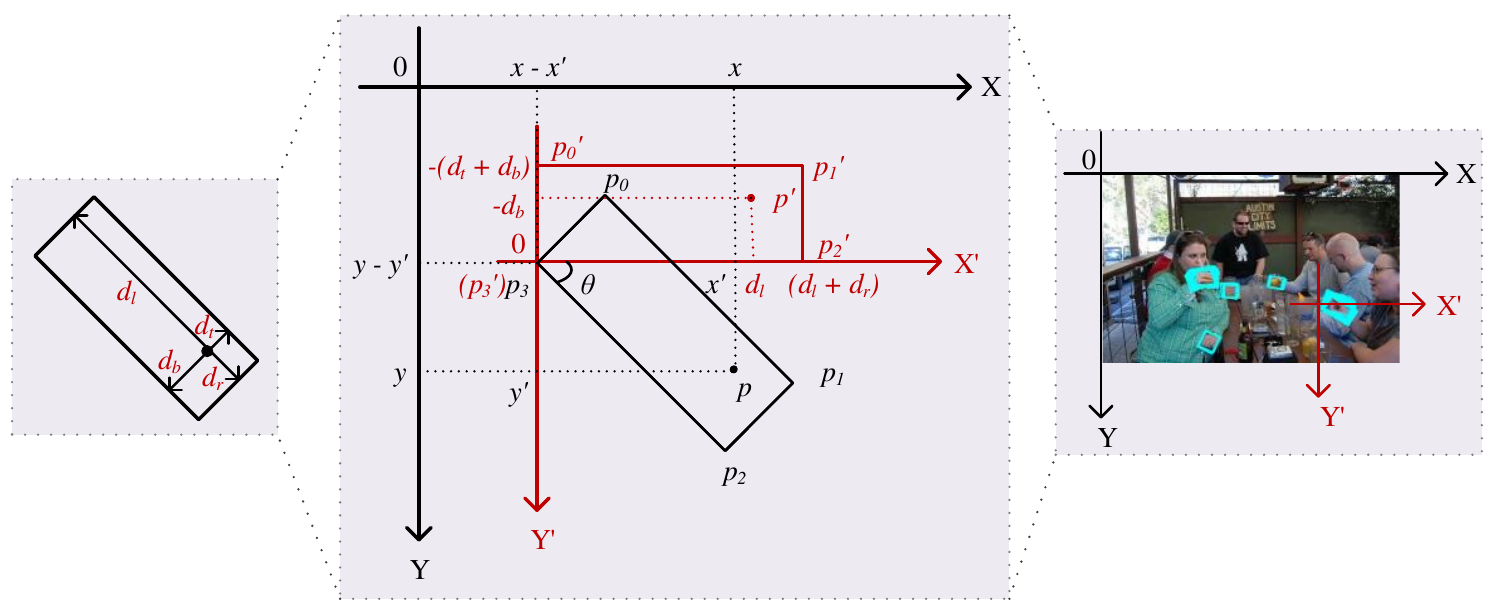}%
	\caption{Restore hand bounding boxes from the rotation map and distance map.}
	\label{restore_rectangle}
\end{figure}
\subsection{Pixel-wise hand detection}
For each pixel in the image, we generate the confidence that it belongs to a hand region and the corresponding hand bounding box. In this way, the model can interpret what features the prediction is based on. The following paragraphs elaborate on this process.

After the last HFF block, the feature maps go through a $3\times3$ convolution and then be upsampled to the same size as the input image. Finally, $1\times1$, $1\times1$ and $3\times3$ convolutions are employed to generate the score map, rotation map and distance map respectively. The three kinds of map are the same size as the original images, and their pixels correspond one by one. Similar to the confidence map used in Fully Convolutional Networks (FCN)~\cite{15}, each pixel value in the score map, a scalar between $0$ and $1$, represents the confidence that the corresponding pixel in the input image belongs to a hand region. The rotation map is developed for the rotated hand detection issue. It records the rotation angle of the hand bounding box and the range of the angle is $(-\pi/2, \pi/2)$. Inspired by the work of \cite{34}, we use the distance map to store the geometry information of the hand box. The distance map has four channels, recording distances to the boundaries of the corresponding hand bounding box, denoted as $d_t,\ d_r,\ d_b,\ d_l$ in Fig.~\ref{restore_rectangle}.

Hand boxes are generated with the rotation map and distance map for pixels whose scores are higher than a given threshold in the score map. An example is given in Fig.~\ref{restore_rectangle} to illustrate the restoring process for pixel $p$. Based on the distance map we can obtain the distances $d_t,\ d_r,\ d_b,\ d_l$ from $p$ to the four boundaries (top, right, bottom, left) of the rectangle $R_p$. In order to calculate the coordinates of $ p_{0},\ p_{1},\ p_{2},\ p_{3}$ in image coordinate system (drawn in black in Fig.~\ref{restore_rectangle}), an auxiliary coordinate system (drawn in red in Fig.~\ref{restore_rectangle}) is introduced with $p_{3}$ as the origin. The directions of X-axis and Y-axis are the same as the image coordinate system. We rotate $R_p$ to the horizontal around $p_{3}$. The corresponding position of $p$ in the rotated rectangle $R'_p$ is denoted as $p'$. Let $(x', y'),\ (x_{i}', y_{i}'),\ i\in\{0, 1, 2\}$ be the coordinates of $p,\ p_i,\ i\in\{0, 1, 2\} $ in the auxiliary coordinate system. For the clockwise rotation of rectangle $R_p$, we have
\begin{equation}
\label{r_pi}
\begin{split}
M\left(\theta \right)
\left(
\begin{array}{c}
x' \\
y'
\end{array}
\right)
&=
\left(
\begin{array}{c}
d_l \\
-d_b
\end{array}
\right),\\
M\left(\theta \right)
\left(
\begin{array}{c}
x_{0}' \\
y_{0}'
\end{array}
\right)
&=
\left(
\begin{array}{c}
0 \\
-(d_t+d_b)
\end{array}
\right), \\
M\left(\theta \right)
\left(
\begin{array}{c}
x_{1}' \\
y_{1}'
\end{array}
\right)
&=
\left(
\begin{array}{c}
d_l+d_r \\
-(d_t+d_b)
\end{array}
\right), \\
M\left(\theta \right)
\left(
\begin{array}{c}
x_{2}' \\
y_{2}'
\end{array}
\right)
&=
\left(
\begin{array}{c}
d_l+d_r \\
0
\end{array}
\right),
\end{split}
\end{equation}
where $M\left(\theta \right)$ is the rotation matrix in two-dimensional space, which can be formulated as
\begin{equation}
\label{r_matrix}
M\left(\theta \right)=
\left(
\begin{array}{cc}
\cos \theta & -\sin \theta \\
\sin \theta & \cos \theta
\end{array}
\right).
\end{equation}
$\theta$ is the rotation angle with counter-clockwise as the positive direction, and it can be restored from the rotation map in our experiments.

Finally, the coordinates $(x_{i}, y_{i}), i\in\{0, 1, 2, 3\}$ of $ p_{i}$ in the image coordinate system are calculated by
\begin{equation}
\label{cal_ords}
\begin{split}
\left(
\begin{array}{c}
x_{3}\\
y_{3}
\end{array}
\right)
&=
\left(
\begin{array}{c}
x\\
y
\end{array}
\right)
-
\left(
\begin{array}{c}
x'\\
y'
\end{array}
\right), \\
\left(
\begin{array}{c}
x_{i}\\
y_{i}
\end{array}
\right)
&=
\left(
\begin{array}{c}
x_{i}'\\
y_{i}'
\end{array}
\right)
+
\left(
\begin{array}{c}
x_{3}\\
y_{3}
\end{array}
\right),\ 
i \in \{0, 1, 2\}.
\end{split}
\end{equation}
$(x, y)$ are the coordinates of $p$ in the image coordinate system. According to Eq.~ \eqref{r_pi}$\sim$\eqref{cal_ords}, the hand bounding box $R_p=\{(x_{i},y_{i})|i\in\{0,1,2,3\}\}$ corresponding pixel $p$ can be restored with the rotation map and distance map.

Many redundant detection bounding boxes are produced by the network. To generate pure detection results, we use the NMS to filter the boxes with low scores and high overlapping rates.
\subsection{Auxiliary Supervision}
The detection loss function usually includes the confidence loss and the location loss. Specific to our method, the confidence loss is calculated with the score map, and the location loss consists the rotation loss and the geometry loss, related to the rotation map and distance map respectively. To learn a more discriminative mask in the HFF, deep supervision is added to the intermediate HFF blocks with auxiliary losses ($L_s,\ s=1,2,3$ in Fig.~\ref{net_arch}) besides the $L_0$ for the output. The overall objective loss function is formulated as
\begin{equation}
\label{loss}
L = \sum\limits_{s \in S}w_{s}L_s,
\end{equation}
where $S=\{0,1,2,3\}$ represents the scale index of the HFF blocks as shown in Fig.~\ref{net_arch} and the parameter $w_{s}$ adjusts the weight of the corresponding scale. For scale $s$, the loss $L_s$ is a weighted sum of the losses for the score map $L^{[s]}_{sco}$, rotation map $L^{[s]}_{rot}$ and distance map $L^{[s]}_{dis}$:
\begin{equation}
\label{scale_loss}
L_s = \alpha L^{[s]}_{sco}+ \beta L^{[s]}_{rot} + L^{[s]}_{dis}.
\end{equation}
The factors $\alpha$ and $\beta$ control the weights of the three loss terms. We describe these three parts of the loss in detail below.

\subsubsection{Loss Function of Score Map}
Regarding the score map as a segmentation of the input image, we use the Dice Similarity Coefficient~\cite{17} (DSC) to construct the loss for score map. DSC measures the similarity between two contour regions. Let $P,\ G$ be the point sets of two contour regions respectively, then the DSC is defined as
\begin{equation}
DSC(P, G) = \dfrac{2 |P\bigcap G|}{|P|+|G|}.
\end{equation}
$|P|$ (\textit{}. $|G|$) represents the number of elements in set $P$ ($G$). As the ground truth of the score map is a binary mask, the dice coefficient can be written as
\begin{equation}
DSC(P, G) = \dfrac {2 \sum \nolimits_{i=1}^N{p_ig_i}}{\sum \nolimits_{i=1}^N{p_i^2}+\sum \nolimits_i^N{g_i^2}},
\end{equation}
where the sums run over all $N$ pixels of the score map. $p_i$ is the the pixel in the score map $P$ generated by the detection network, and $g_i$ is the pixel in the ground truth map $G$. Based on the dice similarity coefficient, the dice loss is proposed and proved to perform well in segmentation tasks~\cite{16,17,18}. Motivated by this strategy, the loss for the score map is formulated as
\begin{equation}
\label{s_loss}
L_{sco} = 1 - \dfrac {2 \sum \nolimits_{i=1}^N{p_ig_i}+\varepsilon_0}{\sum \nolimits_{i=1}^N{p_i^2}+\sum \nolimits_{i=1}^N{g_i^2}+\varepsilon_0},
\end{equation}
where $\varepsilon_0$ is the smooth.

\subsubsection{Loss Function of Rotation Map}
The rotation map stores the predicted rotation angles for corresponding pixels in the input image. The cosine function is adopted to evaluate the distance between the predicted angle $\tilde{\theta}_i$ and the ground truth $\theta_i$. Consequently, we can calculate the loss of rotation map by
\begin{equation}\label{r_loss}
\centering
L_{rot}= 1- \frac{1}{N}\sum_{i=1}^N\cos\left(\tilde{\theta}_i-\theta_i \right).
\end{equation}

\subsubsection{Loss Function of Distance Map}
As for the regression of the object bounding box, the $l_2$ loss~\cite{55} performs the four distances $d_t,\ d_r,\ d_b,\ d_l$ as independent variables, which may mislead the training when only one or two bounds of the predicted box are close to the ground truth. To avoid this, \cite{54} proposes the IoU loss which treats the four distances as a whole. Besides, the IoU loss can handle bounding boxes with various scales as it uses the IoU to norm the four distances to $[0, 1]$. In other words, the IoU loss is scale-invariant, which is important to detect hands of different sizes. The IoU loss for the distance map is calculated as
\begin{equation}\label{d_loss}
\centering
\begin{split}
L_{dis}&=-\frac{1}{N}\sum_{i=1}^N \ln\dfrac{I^{[i]}+\varepsilon_1}{U^{[i]}+\varepsilon_1},\\
I^{[i]} &= I_h^{[i]} * I_w^{[i]},\\
I_h^{[i]} &= min(d_t, \tilde{d_t})+min(d_b, \tilde{d_b}), \\
I_w^{[i]} &= min(d_l, \tilde{d_l})+min(d_r, \tilde{d_r}),\\
U^{[i]} &= X^{[i]} + \tilde{X}^{[i]} - I^{[i]},\\
X^{[i]} &= (d_t + d_b) * (d_l+d_r),\\
\tilde{X}^{[i]} &= (\tilde{d}_t + \tilde{d}_b) * (\tilde{d}_l+\tilde{d}_r),
\end{split}
\end{equation}
where $N$ is the number of pixels in the distance map and $\varepsilon_1$ is the smooth term. $I^{[i]}$ and $U^{[i]}$ denote the intersection and union of the predicted box $\{\tilde{d}_t,\tilde{d}_r, \tilde{d}_b,\tilde{d}_l\}$ and the ground truth $\{d_t,d_r,d_b,d_l\}$ respectively.

\section{Experiments}\label{experiment}
We evaluate our detector on three benchmark datasets: the VIVA hand detection dataset~\cite{11}, the Oxford hand detection dataset~\cite{12} and the VIVA hand tracking dataset~\cite{42}. 

\subsection{Experimental Settings}
All experiments are conducted on an Intel(R) Core(TM) i7-6700K @ 4.00GHz CPU with a single GeForce GTX 1080 GPU. We try two backbone networks: VGG16~\cite{13} and ResNet50~\cite{10} for feature extraction and use the pre-trained models on ImageNet~\cite{20}. We employ the network with the Base Feature Fusion (BFF) block as our base model and conduct ablation experiments to evaluate the performance of the Highlight Feature Fusion (HFF) block and the auxiliary losses.

Training is implemented with a stochastic gradient algorithm using the ADAM scheme. We take the exponential decay learning rate, the initial value of which is $0.0001$ and decays every $10,000$ iterations with rate $0.94$. The weight parameters $w_{s}, s\in\{1,2,3,4\}$ are all set to $1$ for default. The hyper-parameters $\alpha$, $\beta$ are set to $0.01$ and $20$, respectively. Besides, the score map threshold is set to $0.8$. In other words, all the pixels that obtain scores higher than $0.8$ are considered in the bounding box restoration. Then the bounding boxes are filtered by the NMS with a threshold $0.2$.

In order to reduce the over-fitting risk and improve the generalization performance of the model, a variety of data enhancement strategies are employed. We randomly mirror and crop the images, as well as distort the hue, saturation and brightness for color jittering. Due to the limitation of the GPU capacity, the batch size is set as $12$ and all the images are resized to $512\times 512$ before fed into the network in training. When predicting on the test dataset, the original size of the input image is preserved as the network is a fully convolutional network that allows arbitrary sizes of input images.
\begin{table}[!t]
	\renewcommand{\arraystretch}{1.3}
	\caption{Results on VIVA Hand Detection Dataset}
	\label{VIVAResults}
	\resizebox{\textwidth}{!}{
		\begin{tabular}{l lll l}
			\hline
			Methods & \tabincell{c}{Level-1\\(AP/AR)/\%} & \tabincell{c}{Level-2\\(AP/AR)/\%} & Speed/fps & Environment\\
			\hline
			MS-RFCN~\cite{8} & \textbf{95.1/94.5} & 86.0/\textbf{83.4} & 4.65 & \multirow{2}{*}{6 cores@3.5GHz, 32GB RAM, Titan X GPU}\\
			MS-RFCN~\cite{22} & 94.2/91.1 & \textbf{86.9}/77.3 & 4.65 & \\
			Multi-scale fast RCNN~\cite{23} & 92.8/82.8 & 84.7/66.5 & 3.33 & 6 cores@3.5GHz, 64GB RAM, Titan X GPU\\
			FRCNN~\cite{24} & 90.7/55.9 & 86.5/53.3&-&-\\		
			YOLO~\cite{25} & 76.4/46.0 & 69.5/39.1 & 35.00 & 6 cores@3.5GHz, 16GB RAM, Titan X GPU\\
			ACF\_Depth4~\cite{11} & 70.1/53.8 & 60.1/40.4&-&-\\
			\hline
			Ours (VGG16+BFF) & 88.9/82.8 & 72.6/56.7 & 13.88 &\multirow{6}{*}{4 cores@4.0GHz, 32GB RAM, GeForce GTX 1080}\\
			Ours (VGG16+BFF+Auxiliary Losses) & 92.9/88.3 & 80.9/62.7 & 13.16 &\\
			Ours (VGG16+HFF+Auxiliary Losses) & 92.3/89.1 & 83.6/68.8 & 13.10 &\\
			Ours (ResNet50+BFF) & 93.7/89.9 & 83.6/73.6 & 20.40 &\\
			Ours (ResNet50+BFF+Auxiliary Losses) & 94.0/90.1 & 85.7/74.0 & 20.00 &\\
			Ours (ResNet50+HFF+Auxiliary Losses) & \textbf{94.8}/\textbf{91.1} & \textbf{86.3}/\textbf{75.8} & 19.68 &\\
			\hline
	\end{tabular}}
\end{table}

\begin{figure*}[!t]
	\centering
	\includegraphics[width=\textwidth]{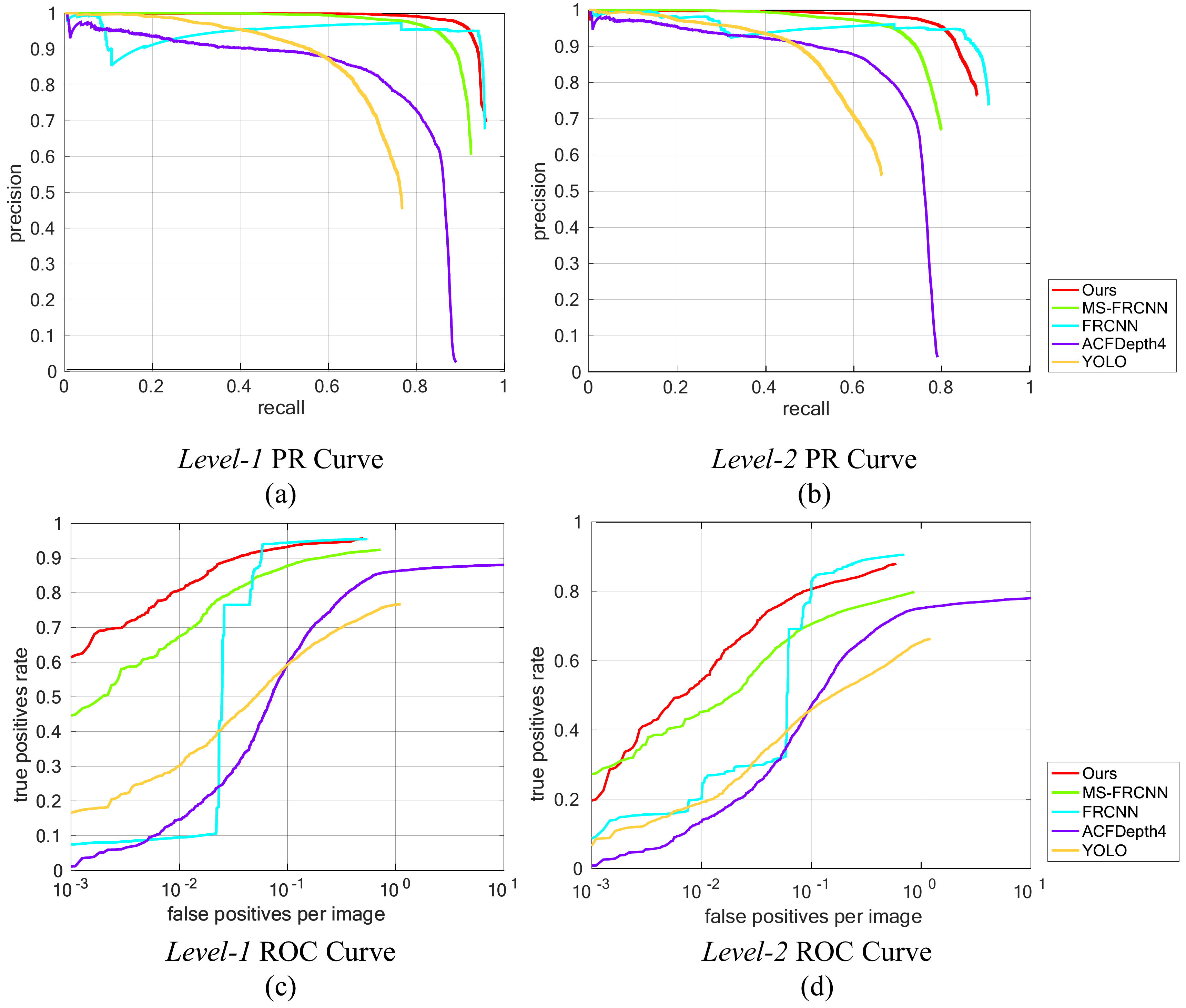}%
	\caption{Precision-Recall curves and ROC curves (logarithmic scale for x-axis) on VIVA dataset.}
	\label{VIVA_curve}
\end{figure*}

\subsection{Evaluations on VIVA Hand Detection Dataset}
VIVA Hand Detection Dataset is published by the Vision for Intelligent Vehicles and Applications Challenge~\cite{11} for hand detection subtask. The dataset includes $5,500$ training and $5,500$ testing images. The images are collected from $54$ videos captured in naturalistic driving scenarios. There are $7$ possible viewpoints in the videos. Annotations for the images are publicly accessible. The bounding boxes of hand regions in an image are given by $(x, y, w, h)$ in the \textit{.txt} format annotation file. $x, y$ are the upper-left coordinates of the box and $w$, $h$ are the width and height of the box, respectively. As the given annotations are axis-aligned, the rotation angles are set to $0$ in training and the predictions are axis-aligned bounding boxes in our experiments on this dataset.

We evaluate the algorithms on two levels according to the size of the hand instances using the evaluate kit provided by the Vision for Intelligent Vehicles and Applications Challenge. \textit{Level-1} focuses on the hand instances with a minimum height of $70$ pixels, only over the shoulder (back) camera view, while \textit{Level-2} evaluates hand samples with a minimum height of $25$ pixels in all camera views. Evaluation metrics include the Average Precision (AP) and Average Recall (AR). AP is the area under the Precision-Recall curve and AR is calculated over $9$ evenly sampled points in log space between $10^{-2}$ and $10^{0}$ false positives per image. As performed in PASCAL VOC~\cite{27}, the hit/miss threshold of the overlap between a pair of predicted and ground truth bounding boxes is set to $0.5$.

As presented in Table.~\ref{VIVAResults}, we compare our methods with MS-RFCN~\cite{8,22}, Multi-scale fast RCNN~\cite{23}, FRCNN~\cite{24}, YOLO~\cite{25} and ACF\_Depth4~\cite{11}. The Precision-Recall curves and ROC curves of these methods and our model (ResNet50+HFF+Auxiliary Losses) are shown in Fig.~\ref{VIVA_curve}. Our model achieves $92.3\%/89.1\%$ (AP/AR) at \textit{Level-1} while $83.6\%/68.8\%$ (AP/AR) at \textit{Level-2} using VGG16 as the backbone network. The ResNet50 based PHDN network obtains more accurate performance, \textit{i.e.}, $94.8\%/91.1\%$ (AP/AR) at \textit{Level-1} and $86.3\%/75.8\%$ (AP/AR) at \textit{Level-2}. 

Apart from the accuracy, the detection speed is also an important metric. As we can see in Table.~\ref{VIVAResults}, YOLO~\cite{25} performs hand detection in real-time, but its accuracy is unsatisfactory. On the contrary, MS-RFCN~\cite{8} performs against other detectors in accuracy but the detecting speed is very slow, \textit{i.e.}, $4.65$ fps. With our PHDN based on VGG16 and ResNet50, the detection speeds are up to $13.10$ and $19.68$ fps, respectively. The model (ResNet50+HFF+Auxiliary Losses) obtains competitive accuracy while a $4.23$ times faster running speed compared to \cite{8}. Therefore, it is of great significance that our model achieves a good trade-off between accuracy and speed.

\begin{table}[!t]
	\renewcommand{\arraystretch}{1.3}
	\caption{Results on Oxford Hand Detection Dataset}
	\label{OxfordResults}
	\centering
	\resizebox{.5\columnwidth}{!}{
		\begin{tabular}{ll}
			\hline
			Methods & AP/\%\\
			\hline
			MS-RFCN~\cite{8} & \textbf{75.1}\\
			Multiple proposals~\cite{12} & 48.2\\
			Multi-scale fast CNN~\cite{23} & 58.4\\
			\hline
			Ours (VGG16+BFF) & 68.7\\
			Ours (VGG16+BFF+Auxiliary Losses) & 77.8\\
			Ours (VGG16+HFF+Auxiliary Losses) & 78.0\\
			Ours (ResNet50+BFF) & 78.2\\
			Ours (ResNet50+BFF+Auxiliary Losses) & 78.6\\
			Ours (ResNet50+HFF+Auxiliary Losses) & \textbf{80.6}\\
			\hline
		\end{tabular}
	}
\end{table}

\begin{figure}[!t]
	\centering
	\includegraphics[width=\textwidth]{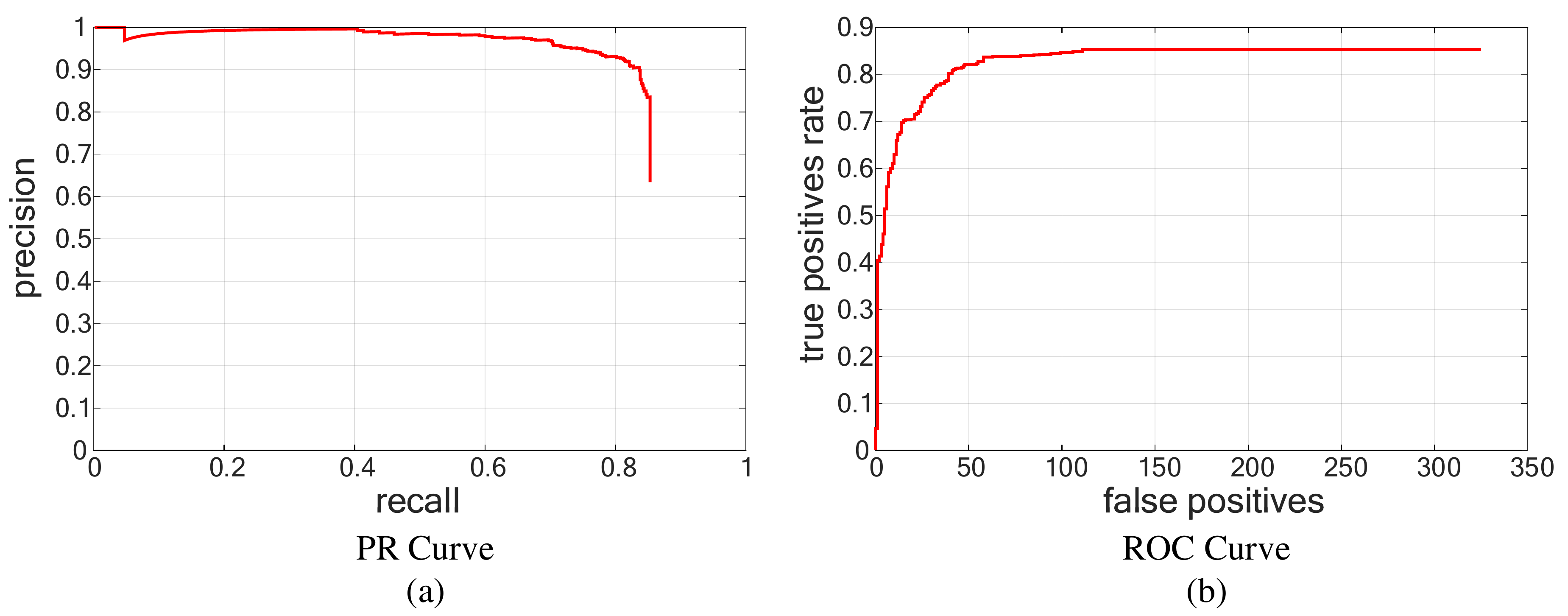}%
	\caption{Precision-Recall curve and ROC curve on oxford dataset.}
	\label{oxford_curve}
\end{figure}

\subsection{Evaluations on Oxford Hand Detection Dataset}
Oxford Hand Detection Dataset consists of three parts: the training set, the validation set and the testing set, with $1,844$, $406$ and $436$ images separately. Unlike the VIVA dataset, the images in Oxford dataset are collected from various different scenes. Moreover, the ground truth is given by the four vertexes $(x_i, y_i), i \in \{1,2,3,4\}$ of the box in the format of \textit{.mat} and not necessarily to be axis-aligned but oriented with respect to the wrist. The rotation angle will be calculated furthermore in our experiments.

According to the official evaluation tool\footnote{\url{http://www.robots.ox.ac.uk/~vgg/data/hands/index.html}} on the Oxford dataset, we report the performance on all the ``bigger'' hand instances, those with more than $1,500$ pixels. As shown in Table.~\ref{OxfordResults}, similar to the results on VIVA dataset, ResNet50 performs better than VGG16 as a backbone network. Specifically, ResNet50 based PHDN achieves an improvement of $5.5\%$ in AP score compared with the state-of-the-art MS-RFCN~\cite{8}. VGG16 based PHDN still outperforms MS-RFCN~\cite{8} by $2.9\%$ in AP score. The Precision-Recall curve and ROC curve are presented in Fig.~\ref{oxford_curve}. In addition, it is worth mentioning that the detecting speed on the Oxford dataset is up to $62.5$ fps using ResNet50 while $52.6$ fps using VGG16.

\begin{table}[!t]
	\renewcommand{\arraystretch}{1.3}
	\caption{Results on VIVA Hand Tracking Dataset}
	\label{VIVATrackingResults}
	\centering
	\resizebox{\textwidth}{!}{
		\begin{tabular}{l|l| llllll ll}
			\hline
			\multicolumn{2}{c|}{Methods} & MOTA/\% & MOTP/\% & Recall/\% & Precision/\% & MT/\% & ML/\% & IDS & FRAG\\
			\hline
			\multirow{4}{*}{Online}&TDC(CNN)~\cite{37} & 25.1 & 64.6 & - & - &39.1 & 18.8 & 34 & 415\\
			& TDC(HOG)~\cite{37} & 24.6 & 64.5 & - & - & 35.9 & 17.2 & 39&426\\
			\cline{2-10}
			& Ours+SORT & 83.4 & \textbf{78.4} & \textbf{90.4} & 92.8 & \textbf{87.5} & 3.13 & 2 & \textbf{88}\\
			& Ours+Deep SORT& \textbf{85.2}& 77.6 & 90.1 & \textbf{94.9} & 84.4 & \textbf{1.56} & \textbf{1} &106\\	
			\hline
			\multirow{2}{*}{Offline}&TBD~\cite{56} & 6.75 & 65.96 & - & - &50 & 12.5 & 29 & 320\\
			\cline{2-10}
			& Ours+IOU & \textbf{83.6} & \textbf{77.1} & \textbf{90.0} & \textbf{93.3} & \textbf{84.4} & \textbf{3.13} & \textbf{5} & \textbf{159}\\
			\hline
		\end{tabular}
	}
\end{table}

\subsection{Evaluations on VIVA Hand Tracking Dataset}
VIVA hand tracking dataset is built by the Vision for Intelligent Vehicles and Applications Challenge for hand tracking sub contest. There are 27 training and 29 test sequences captured under naturalistic driving conditions in this dataset and 2D bounding box annotations of hands are provided with \textit{\{frame, id, bb\_left, bb\_top, bb\_width, bb\_height\}}. Evaluation metrics~\cite{37} follow standard multiple object tracking and are listed as follows.
\begin{itemize}
	\item \textbf{MOTA (The Multiple Object Tracking Accuracy):} A comprehensive metric combining the false negatives, false positives and mismatch rate.
	\item \textbf{MOTP (The Multiple Object Tracking Precision):} Overlap between the estimated positions and the ground truth averaged by all the matches.
	\item \textbf{Recall:} Ratio of correctly matched detections to ground truth detections.
	\item \textbf{Precision:} Ratio of correctly matched detections to total result detections.
	\item \textbf{MT (Most Tracking):} Percentage of ground truth trajectories which are covered by the tracker output for more than 80\% of their length.
	\item \textbf{ML (Most Lost):} Percentage of ground truth trajectories which are covered by the tracker output for less than 20\% of their length.
	\item \textbf{IDS (ID Switches):} Number of times that a tracked trajectory changes its matched ground truth identity.
	\item \textbf{FRAG (Fragments):} Number of times that a ground truth trajectory is interrupted in the tracking result.
\end{itemize}
For MOTA, MOTP, Recall, Precision and MT, greater values mean better performance, whereas the ML, IDS and FRAG are the smaller the better.

To evaluate our detector, we employ the SORT tracker~\cite{39}, deep SORT tracker~\cite{40} and IOU tracker~\cite{41} to associate our detection results to extend a trajectory on the VIVA hand tracking dataset. The results are reported in Table.~\ref{VIVATrackingResults}. The model (ResNet50+HFF+Auxiliary Losses) is used to generate detection results. Note that, we present the Recall and Precision of our method as they are metrics concerned with the detection performance in multiple object tracking. Our model (ResNet50+HFF+Auxiliary Losses) performs much better than the existing methods on this dataset. It indicates that our detector is practicable and well-performed in hand tracking task.

\begin{figure}[!t]
	\centering
	\includegraphics[width=\textwidth]{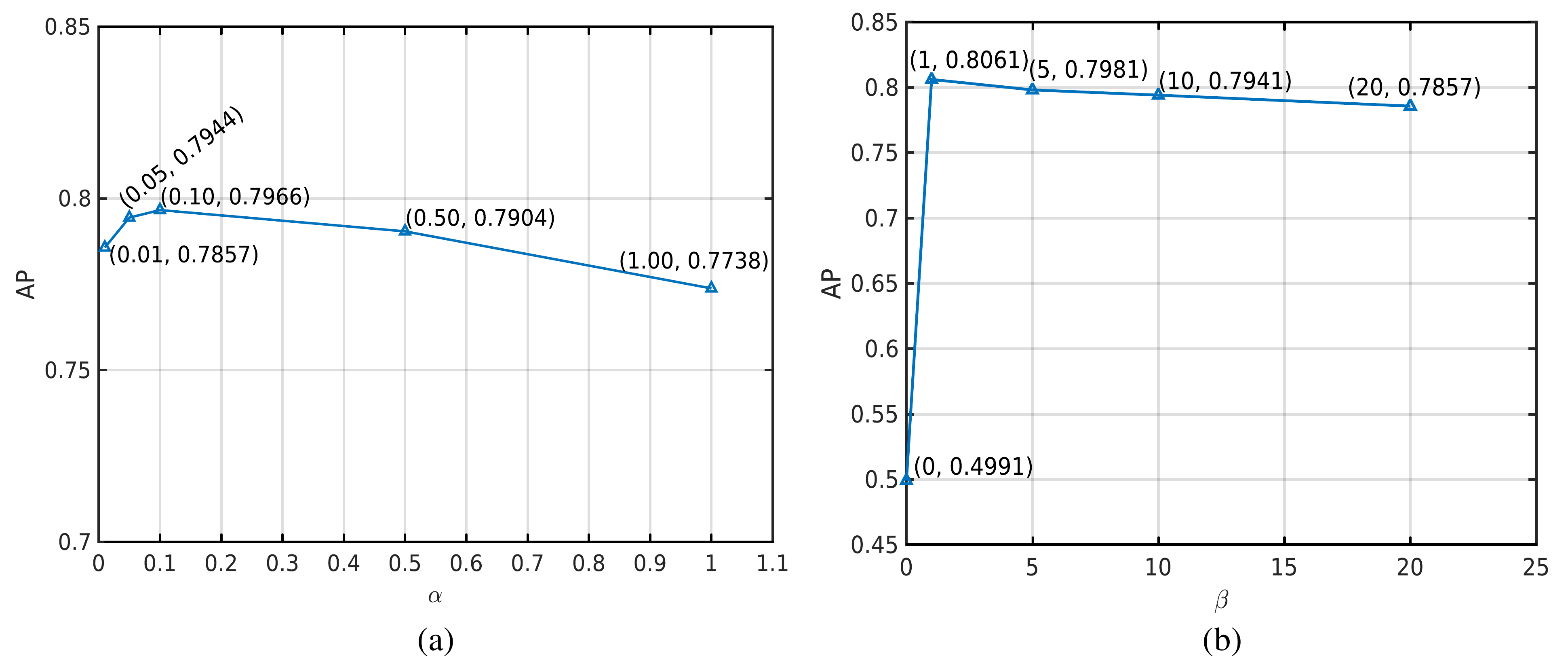}
	\caption{The change of AP with $\alpha$ and $\beta$ on the Oxford dataset.}
	\label{ap}
\end{figure}

\subsection{Ablation Study}
Ablation experiments are conducted to study the effect of different aspects of our model on the detection performance. We choose the ResNet50 as a default backbone network and Oxford hand detection dataset to do further analysis of our model.

\begin{figure}[!t]
	\centering
	\includegraphics[width=4.5in]{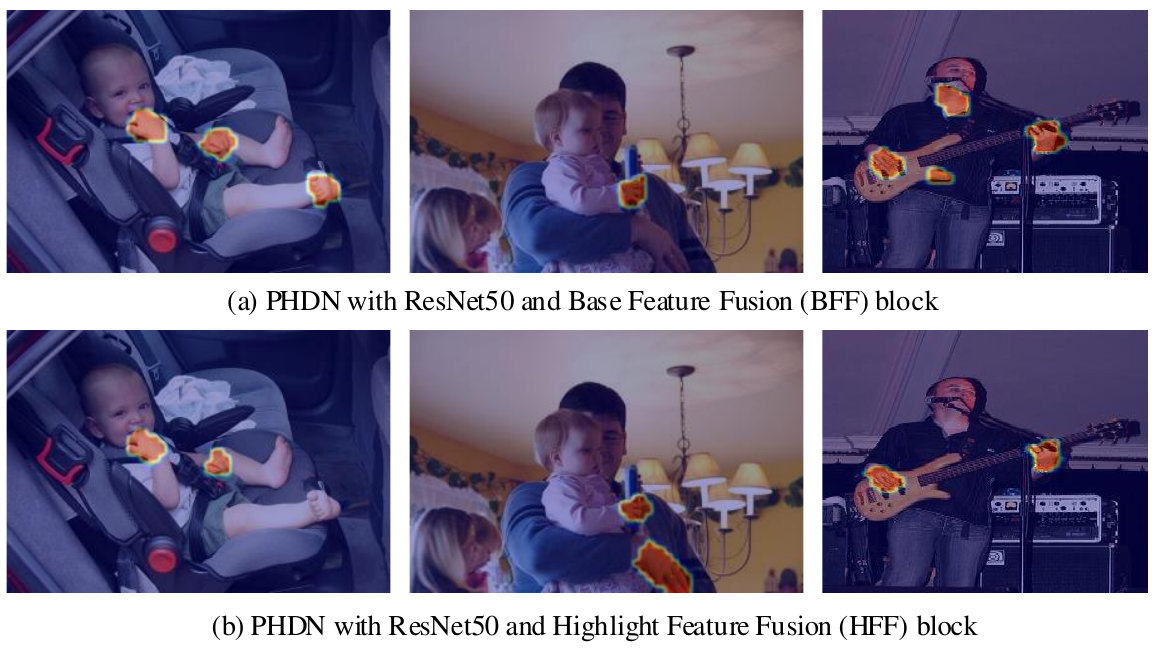}
	\caption{Visual explanations for predictions. The heatmap in the blue-yellow-red color scale is added to the original image to show the activated regions.}
	\label{activation_vis}
\end{figure}

\subsubsection{Interpretable and Robust HFF Block} \label{interpret} Some visual explanations for the effectiveness and robustness of HFF block are given in Fig.~\ref{activation_vis}. The activation feature map is converted into a blue-yellow-red color scale and then added to the original input image to see which pixels are activated in the detection procedure. We can see that the HFF block is good at locating discriminative pixels comparing with the BFF block. The HFF block keeps off confusing parts like faces and feet. It can also activate the hand pixels accurately even in clutter background as shown in the second example in Fig.~\ref{activation_vis}(b). HFF block uses the mask to filter the redundant features of the corresponding layer while the BFF does not. 

From Table.~\ref{VIVAResults} and \ref{OxfordResults}, we can see that the HFF block outperforms the BFF block whether using the VGG16 or ResNet50 as the backbone. Specifically, with VGG16 as the backbone and evaluated at \textit{Level-2}, HFF block achieves an improvement of $2.7\%$ in AP and $ 6.1\%$ in AR on VIVA hand detection dataset. With ResNet50, there are $0.6\%$ in AP and $1.8\%$ in AR respectively. The AR score is improved greatly, which indicates that the model with the HFF block produces less false negatives than the BFF block and makes better use of the distinctive features of different scales. The HFF block also show better performance on the Oxford dataset: It gains an improvement of $0.2\%$ in AP score with VGG16 and $2.0\%$ with ResNet50 comparing to the BFF block.

\subsubsection{Influence of the Score Map and Rotation Map} We adjust the value of $\alpha$ in Eq.~\eqref{loss} to find appropriate weights of score map in training. The results are reported in Fig.~\ref{ap}(a). As $\alpha$ increases from $0.01$ to $1$, the AP increases first and then decreases. It reaches the maximum $0.7966$ when $\alpha$ takes $0.10$ in our experiments. As we can see, if weight the classification loss highly, the AP score will decline ($0.7966$ vs. $ 0.7738 $). In other words, over consideration of score map brings declines in AP score , which is consistent with the fact that the detection is not a simple classification task, but also involves bounding box regression. 

The rotation map is designed to predict the rotation angle of the box and further locate the hand more accurately. To investigate the role it plays in the detection, we control the weights of rotation map in the training process by changing $\beta$ in Eq.~\eqref{loss}. We first set $\beta$ to $0$, \textit{i.e.}, ignore the rotation map in training, to obtain detection results. Then we try four different values ($1$, $5$, $10$ and $20$) for $\beta$ to train models and evaluate all the detection results on the Oxford test set. The AP score and corresponding $\beta$ are plotted in Fig.~\ref{ap}(b) When considering the rotation angle in the optimization procedure, \textit{i.e.}, $\beta>0$, the AP score is stable and larger than $0.78$ for all the values of $\beta$ tried in our experiments. Otherwise, there is a significant drop in the AP score ($0.8061$ vs. $0.4991$) on Oxford dataset when $\beta$ is set as $0$. Therefore, the rotation map plays a very important role in optimizing the final model and can improve the locating accuracy greatly.
\begin{figure}[!t]
	\centering
	\includegraphics[width=0.6\textwidth]{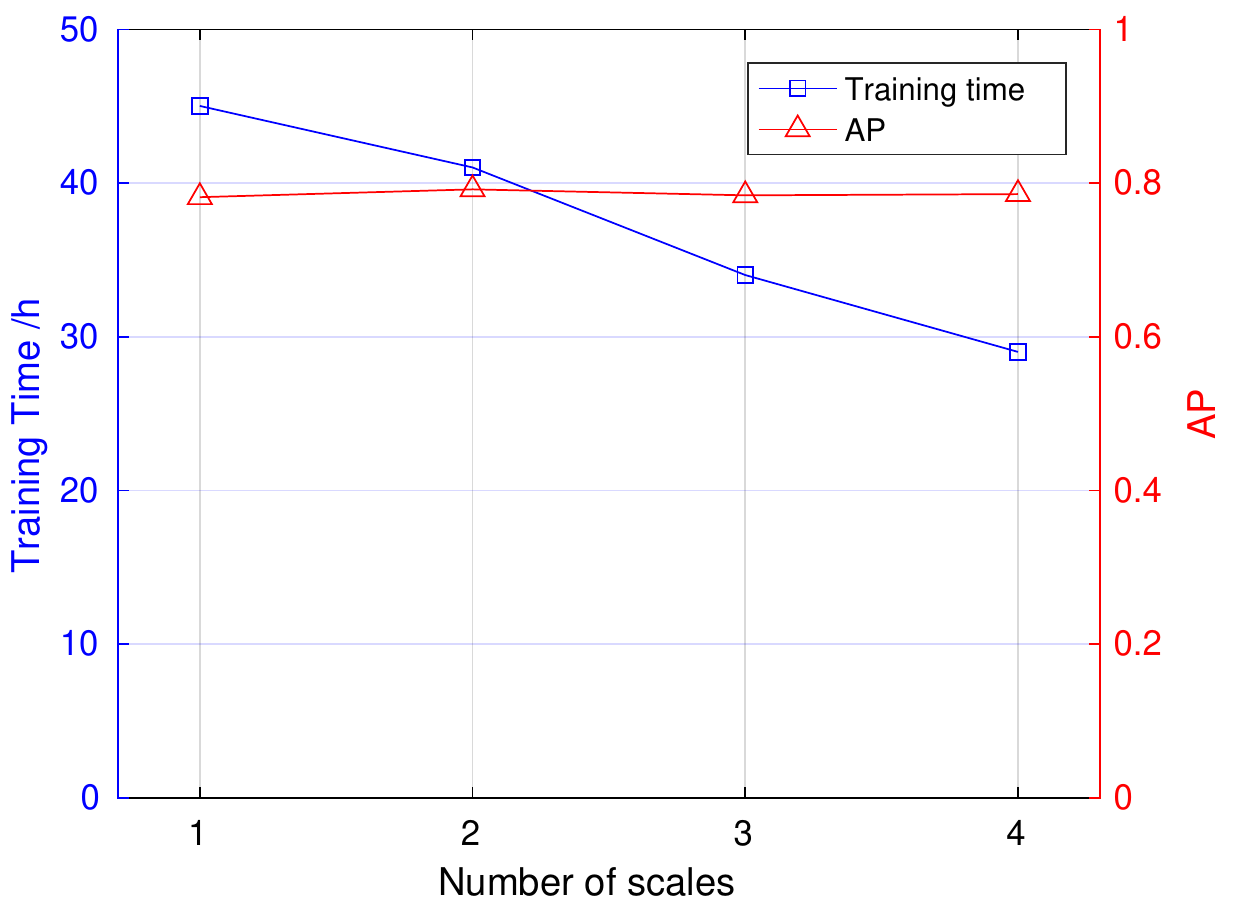}
	\caption{Training time and AP score vs. different numbers of scales on the Oxford dataset.} 
	\label{time}
\end{figure}

\subsubsection{Effectiveness of Auxiliary Supervision} In order to investigate the effectiveness of the auxiliary losses, we train models considering different numbers of scales. The variation of training time and AP score with the number of supervision scales is shown in Fig.~\ref{time}. The number of scales $1, 2, 3, 4$ correspond to $ S=\{0\}, S=\{0,1\}, S=\{0,1,2\}, S=\{0,1,2,3\}$ in Eq.~\eqref{loss} respectively. From Fig.~\ref{time}, we can see that the time it takes for the model to convergence decreases as the number of scales used in loss function increases. The convergence of the network is accelerated significantly (more than $10$ hours) by adding auxiliary losses into the total loss. At the same time, the AP score is stable regardless of the number of scales. It can be concluded that the auxiliary losses accelerate the training process without sacrificing the AP score. This is attributed to the multiple supervision to the intermediate layers of the network.
\begin{figure}[!t]
	\centering
	\includegraphics[width=\textwidth]{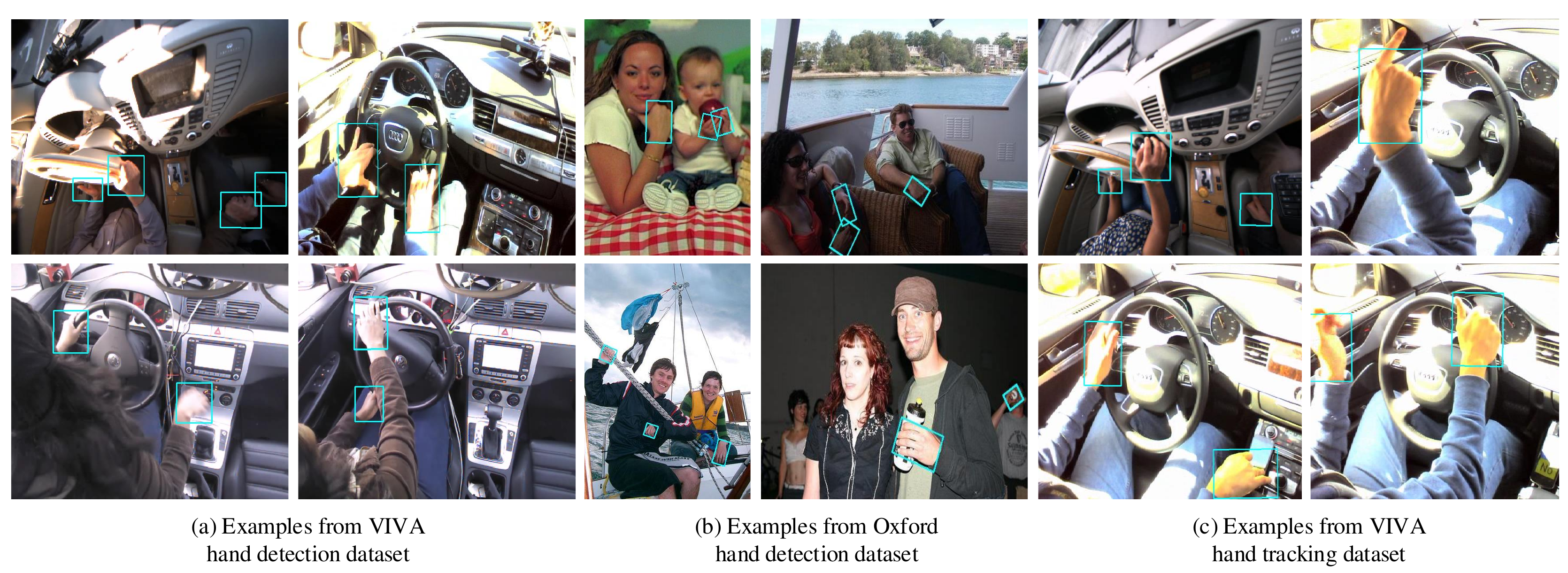}
	\caption{Detection results visualization. Annotations of VIVA hand detection dataset and VIVA hand tracking dataset are horizontal bounding boxes. Images in Oxford hand detection dataset are labeled with wrist-oriented boxes.}
	\label{visualize}
\end{figure}

\begin{figure}[!t]
	\centering
	\includegraphics[width=\textwidth]{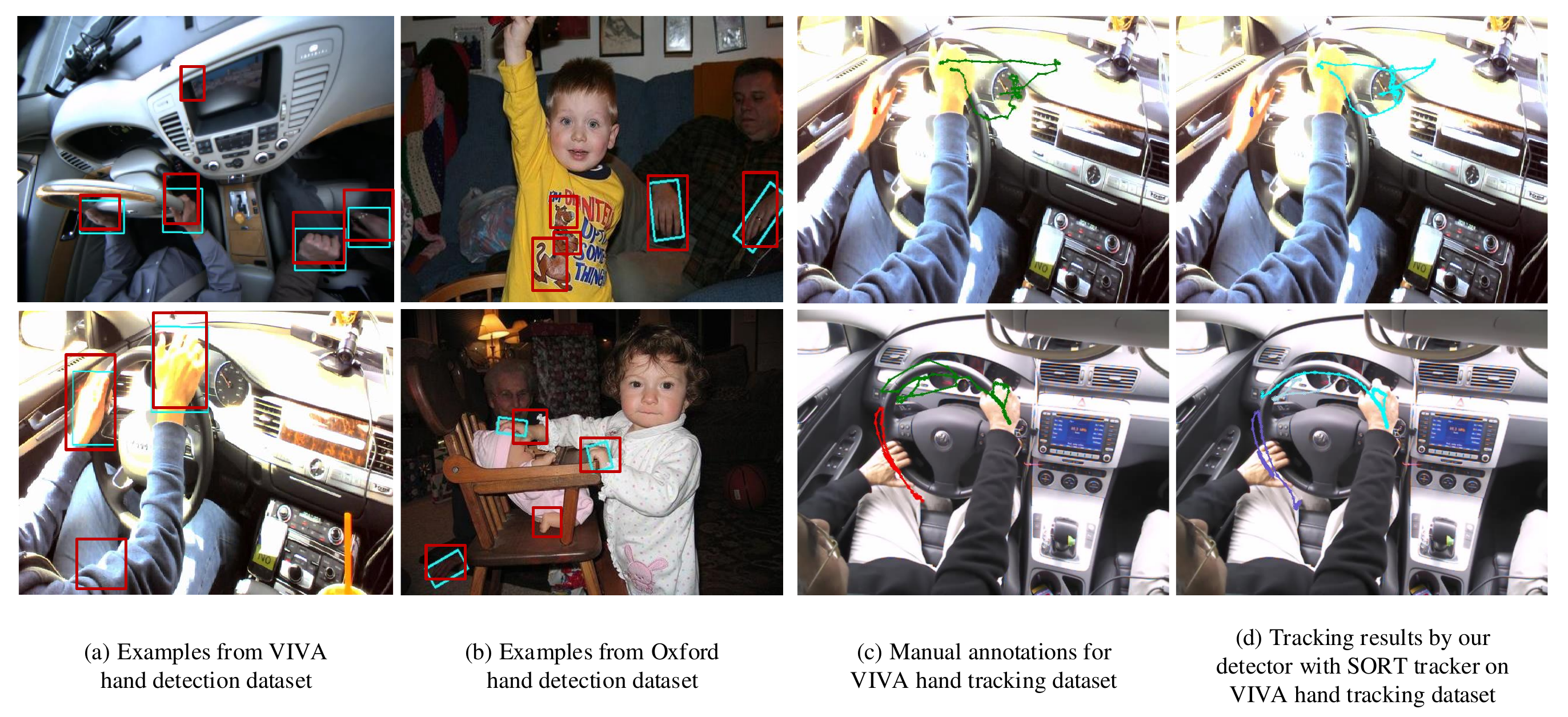}
	\caption{Detection results comparisons. (a) and (b) compare the performance between our PHDN based on ResNet50 model (cyan bounding boxes) and Multi-scale fast RCNN~\cite{23} (red bounding boxes). (c) and (d) show the ground truth and our tracking results on the VIVA hand tracking dataset.}
	\label{comparison}
\end{figure}
\begin{figure}[!t]
	\centering
	\includegraphics[width=0.75\textwidth]{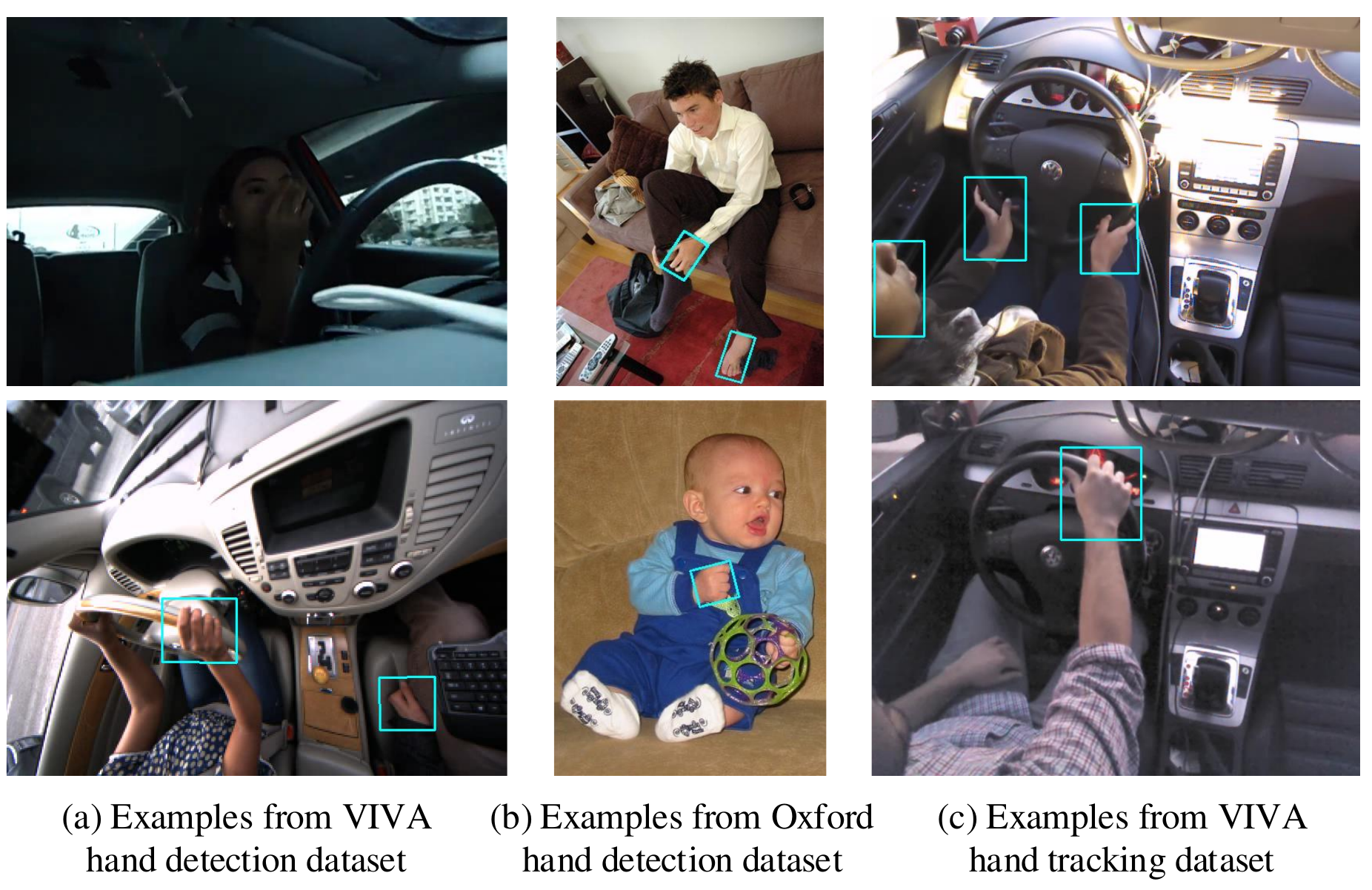}
	\caption{Incorrectly detection examples using PHDN model with ResNet50 as backbone.}
	\label{false}
\end{figure}

\subsubsection{Visualization Results} We show several qualitative detection examples in Fig.~\ref{visualize}. As these results show, our model can handle different scales of hands and shapes in various illumination conditions, even the blurred samples. Fig.~\ref{comparison} compares our detection results with Multi-scale fast RCNN and shows the tracking results and the corresponding ground truth on the VIVA hand tracking dataset. We can see that our model achieves fewer false positives and produces more accurate hand locations compared with the visualization results given in~\cite{23}. Besides, the model trained with rotated hand labels on the Oxford dataset is capable to predict hand rotation angle precisely. Further, applied into the hand tracking task, our model generates satisfactory trajectories as we can see in Fig.~\ref{comparison}. Fig.~\ref{false} shows some false detected samples. The false detections can be divided into three types: (1) When the color or shape of the hand is very close to the background, it may mislead the model to make false predictions or result in missed detection. (2) The faces and feet with confusing colors and shapes are incorrectly detected as hand regions by the model. (3) Heavy occlusions cause missed detection, \textit{e.g.}, the hand obscured by the toy is not recognized in Fig.~\ref{false}(b). Our model does not perform well in these situations possibly because the context information, such as surroundings and similar hand color or shape objects, is not thoroughly mined and integrated effectively. We will investigate the effect of context information in future work and try to address these issues. 

\section{Conclusion}\label{conclusion}
Existing hand detection neural networks are "black box" models and people cannot understand how they make automated predictions. This hinders their application in areas such as driving monitoring.
In this paper, we present the interpretable Pixel-wise Hand Detection Network (PHDN). To the best of our knowledge, this is the first study towards interpretable hand detection. The pixel-wise prediction shows the basis of detection and provides the model interpretability. Features from multiple layers are fused iteratively with cascaded Highlight Feature Fusion (HFF) blocks. This allows our model to learn better representations while reducing computation overhead. The proposed HFF block outperforms the Base Feature Fusion (BFF) block and improves the detection performance significantly. To gain insight into the reasonability of the HFF block, we visualize regions activated by the HFF block and BFF block respectively. The visualization results demonstrate that the HFF block highlights the distinctive features of different scales and learns more discriminative ones to achieve better performance. Complex and non-transparent rotation and derotation layers are replaced by the rotation map to handle the rotated hand samples. The rotation map is interpretable because it directly records the rotation angles of pixels as features. It makes the model more transparent. In addition, deep supervision is added with auxiliary losses to accelerate the training procedure. Compared with the state-of-the-art methods, our algorithm shows competitive accuracy and runs a $4.23$ times faster speed on the VIVA hand detection dataset and achieves an improvement of $5.5\%$ in average precision at a speed of $62.5$ fps on Oxford hand detection dataset. 
Our detector is practical, for which it can track hands better in naturalistic driving conditions compared with other methods on VIVA hand tracking dataset. For future work, we will enhance the transparency and robustness of our model and apply our detector to real-world scenarios such as driving monitoring and virtual reality.

%
%

\section*{References}

\bibliography{myreferences}

\begin{thebibliography}{10}
\expandafter\ifx\csname url\endcsname\relax
  \def\url#1{\texttt{#1}}\fi
\expandafter\ifx\csname urlprefix\endcsname\relax\def\urlprefix{URL }\fi
\expandafter\ifx\csname href\endcsname\relax
  \def\href#1#2{#2} \def\path#1{#1}\fi

\bibitem{zhang2017mdnet}
Z.~Zhang, Y.~Xie, F.~Xing, M.~McGough, L.~Yang, Mdnet: A semantically and
  visually interpretable medical image diagnosis network, in: Proceedings of
  the IEEE conference on computer vision and pattern recognition, 2017, pp.
  6428--6436.
\newblock \href {http://dx.doi.org/10.1109/CVPR.2017.378}
  {\path{doi:10.1109/CVPR.2017.378}}.

\bibitem{montavon2017explaining}
G.~Montavon, S.~Lapuschkin, A.~Binder, W.~Samek, K.-R. M{\"u}ller, Explaining
  nonlinear classification decisions with deep taylor decomposition, Pattern
  Recognition 65 (2017) 211--222.
\newblock \href {http://dx.doi.org/10.1016/j.patcog.2016.11.008}
  {\path{doi:10.1016/j.patcog.2016.11.008}}.

\bibitem{7}
S.~Bambach, S.~Lee, D.~J. Crandall, C.~Yu, Lending a hand: Detecting hands and
  recognizing activities in complex egocentric interactions, in: Proceedings of
  IEEE International Conference on Computer Vision, 2015, pp. 1949--1957.
\newblock \href {http://dx.doi.org/10.1109/ICCV.2015.226}
  {\path{doi:10.1109/ICCV.2015.226}}.

\bibitem{36}
T.~Horberry, J.~Anderson, M.~A. Regan, T.~J. Triggs, J.~Brown, Driver
  distraction: The effects of concurrent in-vehicle tasks, road environment
  complexity and age on driving performance, Accident Analysis \& Prevention
  38~(1) (2006) 185--191.
\newblock \href {http://dx.doi.org/10.1016/j.aap.2005.09.007}
  {\path{doi:10.1016/j.aap.2005.09.007}}.

\bibitem{37}
A.~Rangesh, E.~Ohn-Bar, M.~M. Trivedi, Long-term multi-cue tracking of hands in
  vehicles, IEEE Transactions on Intelligent Transportation Systems 17~(5)
  (2016) 1483--1492.
\newblock \href {http://dx.doi.org/10.1109/TITS.2015.2508722}
  {\path{doi:10.1109/TITS.2015.2508722}}.

\bibitem{3}
P.~Kakumanu, S.~Makrogiannis, N.~Bourbakis, A survey of skin-color modeling and
  detection methods, Pattern Recognition 40~(3) (2007) 1106--1122.
\newblock \href {http://dx.doi.org/10.1016/j.patcog.2006.06.010}
  {\path{doi:10.1016/j.patcog.2006.06.010}}.

\bibitem{2}
A.~Betancourt, P.~Morerio, E.~I. Barakova, L.~Marcenaro, M.~Rauterberg, C.~S.
  Regazzoni, A dynamic approach and a new dataset for hand-detection in first
  person vision, in: Proceedings of International Conference Computer Analysis
  of Images and Patterns, Springer, 2015, pp. 274--287.
\newblock \href {http://dx.doi.org/10.1007/978-3-319-23192-1_23}
  {\path{doi:10.1007/978-3-319-23192-1_23}}.

\bibitem{12}
A.~Mittal, A.~Zisserman, P.~Torr, Hand detection using multiple proposals, in:
  Proceedings of British Machine Vision Conference, 2011, pp. 75.1--75.11.

\bibitem{4}
R.~Girshick, J.~Donahue, T.~Darrell, J.~Malik, Region-based convolutional
  networks for accurate object detection and segmentation, IEEE Transactions on
  Pattern Analysis \& Machine Intelligence 38~(1) (2016) 142--158.
\newblock \href {http://dx.doi.org/10.1109/TPAMI.2015.2437384}
  {\path{doi:10.1109/TPAMI.2015.2437384}}.

\bibitem{6}
W.~Liu, D.~Anguelov, D.~Erhan, C.~Szegedy, S.~Reed, C.~Y. Fu, A.~C. Berg, Ssd:
  Single shot multibox detector, in: Proceedings of European conference on
  computer vision, 2016, pp. 21--37.
\newblock \href {http://dx.doi.org/10.1007/978-3-319-46448-0_2}
  {\path{doi:10.1007/978-3-319-46448-0_2}}.

\bibitem{8}
T.~H.~N. Le, K.~G. Quach, C.~Zhu, N.~D. Chi, K.~Luu, M.~Savvides, Robust hand
  detection and classification in vehicles and in the wild, in: Proceedings of
  IEEE International Conference on Computer Vision \& Pattern Recognition
  Workshops, 2017, pp. 1203--1210.
\newblock \href {http://dx.doi.org/10.1109/CVPRW.2017.159}
  {\path{doi:10.1109/CVPRW.2017.159}}.

\bibitem{23}
S.~Yan, Y.~Xia, J.~S. Smith, W.~Lu, B.~Zhang, Multiscale convolutional neural
  networks for hand detection, Applied Computational Intelligence and Soft
  Computing 2017.

\bibitem{13}
K.~Simonyan, A.~Zisserman, Very deep convolutional networks for large-scale
  image recognition, arXiv preprint arXiv:1409.1556.

\bibitem{10}
K.~He, X.~Zhang, S.~Ren, J.~Sun, Deep residual learning for image recognition,
  in: Proceedings of IEEE International Conference on Computer Vision \&
  Pattern Recognition, 2016, pp. 770--778.
\newblock \href {http://dx.doi.org/10.1109/CVPR.2016.90}
  {\path{doi:10.1109/CVPR.2016.90}}.

\bibitem{9}
X.~Deng, Y.~Yuan, Y.~Zhang, P.~Tan, L.~Chang, S.~Yang, H.~Wang, Joint hand
  detection and rotation estimation by using cnn, IEEE Transactions on Image
  Processing 27~(99).
\newblock \href {http://dx.doi.org/10.1109/TIP.2017.2779600}
  {\path{doi:10.1109/TIP.2017.2779600}}.

\bibitem{60}
L.~Huang, X.~Liu, Y.~Liu, B.~Lang, D.~Tao, Centered weight normalization in
  accelerating training of deep neural networks, in: Proceedings of the IEEE
  International Conference on Computer Vision, 2017, pp. 2803--2811.
\newblock \href {http://dx.doi.org/10.1109/ICCV.2017.305}
  {\path{doi:10.1109/ICCV.2017.305}}.

\bibitem{25}
J.~Redmon, S.~Divvala, R.~Girshick, A.~Farhadi, You only look once: Unified,
  real-time object detection, in: Proceedings of IEEE International Conference
  on Computer Vision \& Pattern Recognition, 2016, pp. 779--788.
\newblock \href {http://dx.doi.org/10.1109/CVPR.2016.91}
  {\path{doi:10.1109/CVPR.2016.91}}.

\bibitem{11}
N.~Das, E.~Ohn-Bar, M.~M. Trivedi, On performance evaluation of driver hand
  detection algorithms: Challenges, dataset, and metrics, in: Proceedings of
  IEEE International Conference on Intelligent Transportation Systems, 2015,
  pp. 2953--2958.
\newblock \href {http://dx.doi.org/10.1109/ITSC.2015.473}
  {\path{doi:10.1109/ITSC.2015.473}}.

\bibitem{42}
\href{http://cvrr.ucsd.edu/vivachallenge/}{Vision for intelligent vehicles and
  applications ({VIVA})}.
\newline\urlprefix\url{http://cvrr.ucsd.edu/vivachallenge/}

\bibitem{39}
A.~Bewley, Z.~Ge, L.~Ott, F.~Ramos, B.~Upcroft, Simple online and realtime
  tracking, in: Proceedings of IEEE International Conference on Image
  Processing, IEEE, 2016, pp. 3464--3468.
\newblock \href {http://dx.doi.org/10.1109/ICIP.2016.7533003}
  {\path{doi:10.1109/ICIP.2016.7533003}}.

\bibitem{40}
N.~Wojke, A.~Bewley, D.~Paulus, Simple online and realtime tracking with a deep
  association metric, in: Proceedings of IEEE International Conference on Image
  Processing, IEEE, 2017, pp. 3645--3649.

\bibitem{41}
E.~Bochinski, V.~Eiselein, T.~Sikora, High-speed tracking-by-detection without
  using image information, in: Proceedings of IEEE International Conference on
  Advanced Video \& Signal Based Surveillance, IEEE, 2017, pp. 1--6.
\newblock \href {http://dx.doi.org/10.1109/AVSS.2017.8078516}
  {\path{doi:10.1109/AVSS.2017.8078516}}.

\bibitem{57}
D.~Liu, D.~Du, L.~Zhang, T.~Luo, Y.~Wu, F.~Huang, S.~Lyu, Scale invariant fully
  convolutional network: Detecting hands efficiently, in: Proceedings of AAAI
  Conference on Artificial Intelligence, 2019.

\bibitem{44}
N.~Dalal, B.~Triggs, Histograms of oriented gradients for human detection, in:
  Proceedings of IEEE International Conference on Computer Vision \& Pattern
  Recognition, Vol.~1, IEEE Computer Society, 2005, pp. 886--893.
\newblock \href {http://dx.doi.org/10.1109/CVPR.2005.177}
  {\path{doi:10.1109/CVPR.2005.177}}.

\bibitem{32}
N.~H. Dardas, N.~D. Georganas, Real-time hand gesture detection and recognition
  using bag-of-features and support vector machine techniques, IEEE
  Transactions on Instrumentation and Measurement 60~(11) (2011) 3592--3607.
\newblock \href {http://dx.doi.org/10.1109/tim.2011.2161140}
  {\path{doi:10.1109/tim.2011.2161140}}.

\bibitem{33}
J.~Niu, X.~Zhao, M.~A.~A. Aziz, J.~Li, K.~Wang, A.~Hao, Human hand detection
  using robust local descriptors, in: Proceedings of IEEE International
  Conference on Multimedia \& Expo Workshops, IEEE, 2013, pp. 1--5.
\newblock \href {http://dx.doi.org/10.1109/ICMEW.2013.6618239}
  {\path{doi:10.1109/ICMEW.2013.6618239}}.

\bibitem{24}
T.~Zhou, P.~J. Pillai, V.~G. Yalla, Hierarchical context-aware hand detection
  algorithm for naturalistic driving, in: Proceedings of IEEE International
  Conference on Intelligent Transportation Systems, 2016, pp. 1291--1297.
\newblock \href {http://dx.doi.org/10.1109/ITSC.2016.7795723}
  {\path{doi:10.1109/ITSC.2016.7795723}}.

\bibitem{15}
J.~Long, E.~Shelhamer, T.~Darrell, Fully convolutional networks for semantic
  segmentation, in: Proceedings of IEEE International Conference on Computer
  Vision \& Pattern Recognition, 2015, pp. 3431--3440.

\bibitem{52}
\href{http://vision.ucsd.edu/pdollar/toolbox/doc/index.html}{P. dollár,
  piotr’s computer vision matlab toolbox (pmt)}.
\newline\urlprefix\url{http://vision.ucsd.edu/pdollar/toolbox/doc/index.html}

\bibitem{20}
A.~Krizhevsky, I.~Sutskever, G.~E. Hinton, Imagenet classification with deep
  convolutional neural networks, in: Proceedings of International Conference on
  Neural Information Processing Systems, 2012, pp. 1097--1105.
\newblock \href {http://dx.doi.org/10.1145/3065386}
  {\path{doi:10.1145/3065386}}.

\bibitem{34}
X.~Zhou, C.~Yao, H.~Wen, Y.~Wang, S.~Zhou, W.~He, J.~Liang, {EAST:} an
  efficient and accurate scene text detector, in: Proceedings of IEEE
  International Conference on Computer Vision \& Pattern Recognition, 2017, pp.
  2642--2651.
\newblock \href {http://dx.doi.org/10.1109/CVPR.2017.283}
  {\path{doi:10.1109/CVPR.2017.283}}.

\bibitem{17}
F.~Milletari, N.~Navab, S.~A. Ahmadi, V-net: Fully convolutional neural
  networks for volumetric medical image segmentation, in: Proceedings of
  International Conference on 3d Vision, 2016, pp. 565--571.
\newblock \href {http://dx.doi.org/10.1109/3DV.2016.79}
  {\path{doi:10.1109/3DV.2016.79}}.

\bibitem{16}
O.~Ronneberger, P.~Fischer, T.~Brox, U-net: Convolutional networks for
  biomedical image segmentation, in: Proceedings of International Conference on
  Medical Image Computing \& Computer-assisted Intervention, 2015, pp.
  234--241.
\newblock \href {http://dx.doi.org/10.1007/978-3-319-24574-4_28}
  {\path{doi:10.1007/978-3-319-24574-4_28}}.

\bibitem{18}
J.~Zhang, X.~Shen, T.~Zhuo, H.~Zhou, Brain tumor segmentation based on refined
  fully convolutional neural networks with a hierarchical dice loss, arXiv
  preprint arXiv:1712.09093.

\bibitem{55}
L.~Huang, Y.~Yang, Y.~Deng, Y.~Yu, Densebox: Unifying landmark localization
  with end to end object detection, arXiv preprint arXiv:1509.04874.

\bibitem{54}
J.~Yu, Y.~Jiang, Z.~Wang, Z.~Cao, T.~Huang, Unitbox: An advanced object
  detection network, in: Proceedings of Acm on Multimedia Conference, ACM,
  2016, pp. 516--520.
\newblock \href {http://dx.doi.org/10.1145/2964284.2967274}
  {\path{doi:10.1145/2964284.2967274}}.

\bibitem{22}
T.~H.~N. Le, C.~Zhu, Y.~Zheng, K.~Luu, M.~Savvides, Robust hand detection in
  vehicles, in: Proceedings of International Conference on Pattern Recognition,
  2017, pp. 573--578.
\newblock \href {http://dx.doi.org/10.1109/ICPR.2016.7899695}
  {\path{doi:10.1109/ICPR.2016.7899695}}.

\bibitem{27}
M.~Everingham, S.~M.~A. Eslami, L.~V. Gool, C.~K.~I. Williams, J.~Winn,
  A.~Zisserman, The pascal visual object classes challenge: A retrospective,
  Int. J. Comput. Vis. 111~(1) (2015) 98--136.
\newblock \href {http://dx.doi.org/10.1007/s11263-014-0733-5}
  {\path{doi:10.1007/s11263-014-0733-5}}.

\bibitem{56}
A.~Geiger, M.~Lauer, C.~Wojek, C.~Stiller, R.~Urtasun, 3d traffic scene
  understanding from movable platforms, IEEE Transactions on Pattern Analysis
  \& Machine Intelligence 36~(5) (2014) 1012--1025.
\newblock \href {http://dx.doi.org/10.1109/tpami.2013.185}
  {\path{doi:10.1109/tpami.2013.185}}.

\end{thebibliography}

\end{document}